\documentclass[lettersize,journal]{IEEEtran}
\usepackage{amsmath,amsfonts}
\usepackage{algorithmic}
\usepackage{algorithm}
\usepackage{array}
\usepackage[caption=false,font=normalsize,labelfont=sf,textfont=sf]{subfig}
\usepackage{textcomp}
\usepackage{stfloats}
\usepackage{url}
\usepackage{verbatim}
\usepackage{graphicx}
\usepackage{cite}
\usepackage{wrapfig}
\usepackage{multirow}
\usepackage{booktabs}

\begin{document}

\title{Robust Motion Averaging for Multi-view Registration of Point Sets Based \\
 Maximum Correntropy Criterion}

\author{Yugeng Huang, Haitao Liu, Tian Huang
\thanks{Yugeng Huang, Haitao Liu and Tian Huang are with the School of Mechanical Engineering, Tianjin University, Tianjin, P.R. China
(e-main: liuht@tju.edu.cn; huangyugeng@tju.edu.cn).}}

\markboth{Journal of \LaTeX\ Class Files,~Vol.~14, No.~8, August~2022}%
{Shell \MakeLowercase{\textit{et al.}}: A Sample Article Using IEEEtran.cls for IEEE Journals}


\maketitle

\begin{abstract}
As an efficient algorithm to solve the multi-view registration problem, 
the motion averaging (MA) algorithm has been extensively studied and many MA-based 
algorithms have been introduced. They aim at recovering global motions from relative 
motions and exploiting information redundancy to average accumulative errors. However, 
one property of these methods is that they use Guass-Newton method to solve a least squares 
problem for the increment of global motions, which may lead to low efficiency and 
poor robustness to outliers. In this paper, we propose a novel motion averaging 
framework for the multi-view registration with Laplacian kernel-based maximum 
correntropy criterion (LMCC). Utilizing the Lie algebra motion framework and the 
correntropy measure, we propose a new cost function that takes all constraints 
supplied by relative motions into account. Obtaining the increment used to 
correct the global motions, can further be formulated as an optimization problem 
aimed at maximizing the cost function. 
By virtue of the quadratic technique, the optimization problem 
can be solved by dividing into two subproblems, i.e., computing the weight for 
each relative motion according to the current residuals and solving a second-order 
cone program problem (SOCP) for the increment in the next iteration. We also provide 
a novel strategy for determining the kernel width which ensures that our method 
can efficiently exploit information redundancy supplied by relative motions 
in the presence of many outliers. Finally, we compare the proposed method with 
other MA-based multi-view registration methods to verify its performance. 
Experimental tests on synthetic and real data demonstrate that our method achieves 
superior performance in terms of efficiency, accuracy and robustness. 
\end{abstract}


\section{Introduction}
\IEEEPARstart{P}{oint} set registration is a fundamental but challenging methodology, 
which is widely used in the fields of computer vision, computer graphics, and robotics,
 for 3D model reconstruction, simultaneous localization and mapping (SLAM), 
 cultural heritage management, etc. With the development of 3D scanning devices, 
 it has become easy to obtain point cloud representations of real-world models or 
 scenes. However, a non-negligible fact is that due to the occlusion of scenes or objects 
 and the limited scanning field of view, most scanning devices only capture partial point 
 clouds at one time. Thus, in order to accurately recover the whole model or scene from a 
 sequence of scans captured from different viewpoints, point set registration problem 
 arises. It aims at determining the rigid transformations of these scans between a given 
 centered-frame so as to align multi-view scans into an entire model or scene. Depending 
 on the amount of input point clouds, point set registration can be divided into two 
 problems, namely, pair-wise registration and multi-view registration \cite{ref1}. 

 In the past few decades, several methods have been proposed to solve the pair-wise registration problem 
 and these methods can be classified into two categories, i.e., coarse registration methods and fine registration 
 methods \cite{ref1}. For the former, the feature-based methods are the most popular and have been extensively studied, 
 which usually contain three significant steps: extracting geometric features composed of keypoints and 
 corresponding descriptors, matching features to obtain the correspondences, and estimating the rigid transformation 
 based on the correspondences. Usually, the performance of the feature-based methods is affected by point density, 
 outliers, and overlapping percentage \cite{ref2}. The deep learning-based methods, widely studied in recent years, are 
 likely to overcome these drawbacks \cite{ref1}. Compared to coarse registration, fine registration can achieve more 
 accurate results for pair-wise registration. The ICP-based algorithms, 
 among these fine registration methods, are the ones studied most intensively. By alternately searching the nearest correspondences from the target points for each 
 source point and estimating and updating the rigid transformation until convergence, the ICP-based algorithms achieve an accurate registration result \cite{ref3}. 
 Since the original ICP algorithm \cite{ref3} minimizes the squared distances of correspondences, it is not robust to outliers and 
 ignores the structural information of the point set. Thus, many variants \cite{ref4,ref5,ref6} of ICP are proposed to improve 
 the robustness to outliers, efficiency, and accuracy. For example, 
 point-to-plane metric \cite{ref5}, plane-to-plane metric \cite{ref6}, 
 and symmetric point-to-plane metric \cite{ref7} are proposed 
 to take advantage of the structural information. In addition, good initializations are necessary for ICP-based methods 
 to guarantee efficiency and the avoidance of local minima.

 Compared to pair-wise registration, multi-view registration is more difficult due to many parameters needed to be determined. 
 For solving multi-view registration problems, the most common methods are sequential pair-wise registration-based methods and 
 joint registration-based methods [1]. For the former, a registration-and-integration strategy is widely adopted and the pair-wise registration 
 algorithms play important roles in the registration process. For example, Chen et al. \cite{ref9} repeatedly registered 
 two scans with the utilization of ICP and integrated a new scan until all of the scans are translated into the whole model. In addition, treating the 
 multi-view registration problem as a graph optimization issue, Weber et al. [10] estimated the overlap of each scan pair as the weight of the 
 corresponding edge so as to establish a weighted graph with fully connected edges. Then, a minimum 
 spanning tree (MST) can be constructed by utilizing the Kruskal algorithm \cite{ref11}, and finally, the rigid transformation of each node can be 
 sequentially estimated by performing a breadth-depth search on the resulting MST. Furthermore, Zhu et al. \cite{ref12} proposed a dual criterion 
 to judge the reliability of pair-wise registration results and picked out the relatively reliable pairs to construct the spanning tree. 
 Although sequential pair-wise registration-based methods are efficient, they neglect the relevance of each scan and do not exploit 
 information redundancy in the point set, resulting in the accumulated error.

 To reduce the error, joint registration-based methods have been developed in recent years. These methods fall into two main categories, i.e., the expectation maximization (EM)-based 
 algorithms \cite{ref13}\cite{ref14} and motion averaging (MA)-based algorithms \cite{ref8}\cite{ref15,ref16,ref17}. The former aligns all scans based on probabilistic models, 
while the latter recovers global 
 motions from a set of relative motions so as to exploit information redundancy and further average the registration errors. In addition, to the 
 best of our knowledge, these methods share a common drawback that they are local convergent algorithms and a good initialization is necessary 
 for them to achieve accurate registration results. By virtue of the Lei algebraic property of the rigid-body motion, Govindu 
 et al. \cite{ref15} first proposed the motion averaging (MA) method. MA takes an initial global motions and a set of 
 relative motions of scan pairs as input. Different from the sequential pair-wise registration-based methods, MA takes advantage of information 
 redundancy in relative motions. By utilizing a $\ell_{2}$-based optimization considering all relative motions, MA can efficiently average the accumulative errors to each scan. 
 However, $\ell_{2}$-based measure is sensitive to outliers, resulting in poor robustness. To improve the robustness, 
  Govindu \cite{ref18} further treated the outlier rejection problem as a graph optimization issue. Although the proposed method eliminates the effect of outliers, 
  the Random Sample Consensus introduced in this method is very time-consuming. 
  To reduce the outliers ratio of relative motions, Li et al. \cite{ref16} 
 proposed an improved technique for MA-based multi-view registration, known as MATrICP. In this technique, the overlapping percentage 
 of each scan pair is roughly estimated, and then these scan pairs whose overlapping percentages are larger than a pre-set threshold 
 are picked out. Finally, MATrICP applies TrICP algorithm to obtain the relative motions corresponding to these scan pairs as the 
 input of MA.
  Inspired by MATrICP, Guo et al. \cite{ref17} proposed a weighted motion averaging (WMA) algorithm for multi-view registration, called WMAA.
  Considering the reliability and accuracy of each relative motion, WMAA takes the overlapping percentage determined by TrICP algorithm, 
  as the corresponding weight of each scan pair. WMAA is more likely to return 
 desired registration results than MA \cite{ref8}. 
 However, similar to most of the variants of ICP, the accuracy of TrICP algorithm is also highly dependent on good initializations and 
 easily traps into local minima. 
 In addition, Arrigoni et al. \cite{ref19} introduced the low-rank and sparse (LRS) matrix decomposition to recover global 
 motions from a set of relative motions, and therefore, LRS can be cast as another MA algorithm. 
 Although LRS achieves superior performance in terms of efficiency in MA-based multi-view registration, the performance of LRS is more 
 susceptible to the sparsity of view-graph than related methods \cite{ref20}. 
 Additionally, assuming all the points are drawn from a central Gaussian mixture model (GMM), 
 Evangelidis and Horaud \cite{ref13} proposed an EM-based method, called JRMPC, which casts multi-view registration into a clustering 
 problem.
  Furthermore, Zhu et al. \cite{ref14} proposed the EMPMR method by assuming that each data point is generated from one unique GMM, where 
 its corresponding points in other point sets are regarded as Gaussian centroids. 

 In recent years, a method, called correntropy, is introduced to improve the robustness to outliers in information-theoretic learning \cite{ref21}\cite{ref22}. 
 Nowadays, the method is also introduced 
 to the field of computer vision, for example, based on MCC, He et al. \cite{ref23} proposed a correntropy framework for robust face recognition. Zhu 
 et al. \cite{ref8} further combined MCC with the motion averaging algorithm, and proposed a robust framework for multi-view registration (MCC-MA). Specifically, by introducing Gaussian kernel-based MCC, MCC-MA reformulates the MA as a half-quadratic optimization problem that can be readily solved by the traditional 
 iterative method introduced in \cite{ref15,ref16,ref17}. MCC-MA has achieved state-of-the-art 
 performance in terms of accuracy in multi-views registration based MA framework. However, its runtime rapidly increases with the increase in the number of relative motions. 

 Inspired by MCC-MA, a novel and iterative motion averaging framework is proposed in this paper to 
 further enhance the efficiency and robustness of MA-based multi-view registration.
 Different from the Gaussian kernel introduced in \cite{ref8}, we adopt the Laplacian kernel in MCC and 
 provide a robust strategy for determining the kernel width.  
 Specifically, by utilizing the Lie algebra motions framework and the correntropy measure, we propose 
 a new cost function that takes all constraints supplied by relative motions into account. 
 The unknowns, i.e., the increment of global motions in Lie algebra form, can be solved by maximizing 
 the cost function. An approach to the solution of this maximization problem consists in two steps: 
 In the first step, the weights for each pair are calculated according to the current residuals, 
 thereby under the characteristic of Laplacian function, these pairs whose corresponding relative 
 motions are reliable will be assigned large weights and vice versa; In the second step, decoupled 
 from the nonlinearity of Laplacian function by introducing additional variables, solving for the
  unknowns is readily done by solving a second-order cone program problem (SOCP). The relevance of 
  these two problems in connection with the performance of our method is needed to be stressed: the 
  first step allows us to efficiently eliminate the effect of outliers, while the second step not 
  only avoids the calculation of the pseudo-inverse of a large matrix which is a fundamental and 
  unavoidable step in Gauss-Newton method adopted in MA \cite{ref15}, WMAA \cite{ref17} and MCC-MA \cite{ref8}, but also reduces 
  the number of iterations required for convergence due to the high speed of convergence of SOCP. 

In summary, the contributions of our works are described as follows:

1) We elaborate further on the motion averaging algorithm in multi-view registration (Section II) and propose a novel MA-based algorithm, called L-MA. We further extend the concept of MCC to multi-view registration by combing Laplacian kernel-based MCC with motion averaging (Section III). Experimental tests on synthetic and real data demonstrate that our method achieves state-of-the-art performance (Section IV).  

2) We propose a new cost function for motion averaging. We solve an optimization problem for the increment of global motions by utilizing a half-quadratic technique introduced in \cite{ref24} (Section III). 

3) We discuss the shortcomings of most existing MA-based methods for multi-view registration (Section II), and we explain why our method can efficiently overcome these drawbacks (Section III).

\section{Problem Formulation and Weighted Motion Averaging Algorithm Revisited}

Given a data set containing $n$ scans, the rigid transformation of the $i$-th scan to the set-centered frame ${{\mathcal{K}}_{w}}$ can be denoted by 

\begin{equation}
\label{deqn_ex1a}
{{\mathbf{M}}_{i}}=\left[ \begin{matrix}
  {{\mathbf{R}}_{i}} & {{{\boldsymbol{t}}}_{i}}  \\
  {{\mathbf{0}}^{\text{T}}} & 1  \\
\end{matrix} \right],{{\mathbf{M}}_{i}}\in SE(3)
\end{equation}
where ${{\mathbf{R}}_{i}}\in SO(3)$ and ${{\boldsymbol{t}}_{i}}\in {{\mathbb{R}}^{3}}$ denote the rotation matrix and the translation vector, respectively. Furthermore, for ${{\mathbf{M}}}\in SE(3)$, there exists a corresponding vector $\boldsymbol{\xi} \in {{\mathbb{R}}^{6}}$. $\boldsymbol{\xi} \in {{\mathbb{R}}^{6}}$ can be converted from vector space to Lie algebra space 
$\mathfrak{s}\mathfrak{e}(3)$ by ${\left(\cdot \right)}^{\wedge}$ operation, i.e., $\boldsymbol{\xi} \in {{\mathbb{R}}^{6}}\to {{\boldsymbol{\xi} }^{\wedge }}\in {{\mathbb{R}}^{4\times 4}}$

\begin{equation}
\label{deqn_ex2a}
{{\boldsymbol{\xi }}^{\wedge }}=\left[ \begin{matrix}
  {{\left[ \boldsymbol{\theta } \right]}_{\times }} & \boldsymbol{\rho }  \\
  {{\mathbf{0}}^{\text{T}}} & 0  \\
\end{matrix} \right],\boldsymbol{\xi }=\left[ \begin{matrix}
  \boldsymbol{\rho }  \\
  \boldsymbol{\theta }  \\
\end{matrix} \right]
\end{equation}
where $\boldsymbol{\theta} \in {{\mathbb{R}}^{3}}$ and $\boldsymbol{\rho} \in {{\mathbb{R}}^{3}}$ and are the rotational and translational components of $\boldsymbol{\xi}$, 
respectively; ${{\left[ \boldsymbol{\theta } \right]}_{\times }}$ is the corresponding skew-symmetric matrix of $\boldsymbol{\theta} \in {{\mathbb{R}}^{3}}$. Here, we use 
${{\left( \cdot  \right)}^{\vee }}$ operation to convert ${{\boldsymbol{\xi} }^{\wedge }}\in \mathfrak{s}\mathfrak{e}(3)$ to vector space, that is, ${{\left( {{\boldsymbol{\xi} }^{\wedge }} \right)}^{\vee }}\in {{\mathbb{R}}^{4\times 4}}\to \boldsymbol{\xi} \in {{\mathbb{R}}^{6}}$. 
In addition, the relations between ${{\mathbf{M}}}\in SE(3)$ and its corresponding Lie algebra ${{\boldsymbol{\xi} }^{\wedge }}\in \mathfrak{s}\mathfrak{e}(3)$, 
given by the exponential map $\exp (\cdot )$ and its inverse map $\log (\cdot )$, take on the forms 
\begin{equation}
\label{deqn_ex3a}
\mathbf{M} =\exp \left( {{\boldsymbol{\xi} }^{\wedge }} \right),{{\boldsymbol{\xi} }^{\wedge }}=\log \left(\mathbf{M}  \right)
\end{equation}

Without loss of generality, let the first scan be the reference scan, i.e., ${{\mathbf{M}}_{1}}=\mathbf{I}$, where $\mathbf{I}\in {{\mathbb{R}}^{4\times 4}}$ is an identity matrix. Therefore, the goal of multi-view registration is to estimate 
the set of all global motions denoted by ${{\mathbf{M}}_{G}}=\left\{ {{\mathbf{M}}_{i}} \right\}_{i=1}^{n}$, so as to transform scans from local frames into the given set-centered frame $\mathbf{I}\in {{\mathbb{R}}^{4\times 4}}$. 
To this purpose, WMA or/and MA recover global motions from a given set of relative motions $\left\{ {{\mathbf{M}}_{ij,h}} \right\}_{h=1}^{m}$, 
where the subscript $h$ stands for the $h$-th pair of $\left\{ {{\mathbf{M}}_{ij,h}} \right\}_{h=1}^{m}$; ${{\mathbf{M}}_{ij}}$ denotes the motion that aligns 
$i$-th scan to $j$-th scan and can be estimated by utilizing pair-wise algorithm, such as ICP \cite{ref3} and its variants \cite{ref7}\cite{ref25}\cite{ref26}. 

Motivated by the early and excellent works done by Govindu et al. \cite{ref15} and Zhu et al. \cite{ref8} , we will give a more detailed derivation of WMA algorithm than those derivations given in \cite{ref8}\cite{ref15}\cite{ref17} which lies at the basis of our proposed method, and then discuss the shortcomings of previous MA-based methods, as follows:

Ideally, given the relative motion ${{\mathbf{M}}_{ij}}$, the following holds
\begin{equation}
\label{deqn_ex4a}
{{\mathbf{M}}_{ij}}={{\mathbf{M}}_{i}}\mathbf{M}_{j}^{-1},\forall \left( i,j \right)\in E
\end{equation}
\begin{equation}
\label{deqn_ex5a}
\mathbf{M}_{i}^{-1}{{\mathbf{M}}_{ij}}\mathbf{M}_{j}^{{}}=\mathbf{I},\forall \left( i,j \right)\in E
\end{equation}
where $E$  denotes the view-graph whose edge $\left( i,j \right)$ denotes ${{\mathbf{M}}_{ij}}$ and node $i$ denotes ${\mathbf{M}}_{i}$. Unfortunately, when there is no or very low overlap area 
between two scans, it is difficult or even impossible to obtain the accurate relative motion between them. Furthermore, the accuracy 
of the pair-view registration algorithm is usually affected by the initial value and local convergence. Thus, the noise will be such that the constraint equations (4) and (5) will never be 
exactly satisfied. We therefore have 
\begin{equation}
\label{deqn_ex6a}
\mathbf{M}_{i}^{-1}{{\mathbf{M}}_{ij}}\mathbf{M}_{j}^{{}} \approx \mathbf{I},\forall \left( i,j \right)\in E
\end{equation}

In view of (6) and under the condition that the initial global motions $\mathbf{M}_{G}^{0}=\left\{ \mathbf{M}_{i}^{0} \right\}_{i=1}^{n}$ is given, the following notations are adopted: 
(a) let $\mathbf{M}_{G}^{k-1}\text{=}\left\{ \mathbf{M}_{i}^{k-1} \right\}_{i=1}^{n}$ 
be the set of estimated global motion in $(k-1)$-th iteration; 
(b) the increment of the $\mathbf{M}_{G}^{k-1}$ in the $k$-th iteration, which is used to correct $\mathbf{M}_{G}^{k-1}$, is denoted by 
$\Delta \mathbf{M}_{G}^{k}=\left\{ \Delta \mathbf{M}_{i}^{k} \right\}_{i=1}^{n}$; 
(c) we define $\Delta \mathbf{M}_{ij}^{k}={{\left( \mathbf{M}_{i}^{k} \right)}^{-1}}{{\mathbf{M}}_{ij}}\mathbf{M}_{j}^{k}$ as the residual term.
Once $\Delta \mathbf{M}_{G}^{k}$ is estimated, we can obtain the global motion in the next iteration upon multiplication of each global motion 
$\mathbf{M}_{i}^{k-1}$ from the right by the “small” increment $\Delta \mathbf{M}_{i}^{k}$, which is denoted by $\mathbf{M}_{G}^{k}\text{=}\left\{ \mathbf{M}_{i}^{k-1}\Delta \mathbf{M}_{i}^{k} \right\}_{i=1}^{n}$. 
In addition, $\Delta \mathbf{M}_{ij}^{k}$ can be considered as the inconsistency between the global motions 
$\mathbf{M}_{i}^{k}$, $\mathbf{M}_{j}^{k}$ and relative motion ${{\mathbf{M}}_{ij}}$.

With the notations introduced above, the following always holds:
\begin{equation}
\label{deqn_ex7a}
\Delta \mathbf{M}_{ij}^{k}={{\left( \Delta \mathbf{M}_{i}^{k} \right)}^{-1}}\Delta \mathbf{M}_{ij}^{k-1}\Delta \mathbf{M}_{j}^{k}
\end{equation}

For the sake of conciseness, we denote here the corresponding Lie algebra of $\Delta \mathbf{M}_{i}^{k}$,  $\Delta \mathbf{M}_{ij}^{k}$ by $\Delta \boldsymbol{\xi}_{i}^{k}$,  $\boldsymbol{\xi}_{ij}^{k}$, respectively. Thus, the relationship between
$\Delta \mathbf{M}_{i}^{k}$ and $\Delta \mathbf{M}_{ij}^{k}$ of (7) can then be written in Lie algebra form as 
\begin{equation}
\label{deqn_ex8a}
\boldsymbol{\xi} _{ij}^{k}=\log {{\left( \exp {{\left( -\Delta \boldsymbol{\xi} _{i}^{k} \right)}^{\wedge }}\exp {{\left( \boldsymbol{\xi} _{ij}^{k-1} \right)}^{\wedge }}\text{exp}{{\left( \Delta \boldsymbol{\xi} _{j}^{k} \right)}^{\wedge }} \right)}^{\vee }}
\end{equation}

Furthermore, equation (8) shows that $\boldsymbol{\xi} _{ij}^{k}$ is a nonlinear function of $\Delta \boldsymbol{\xi} _{i}^{k}$, and of $\Delta \boldsymbol{\xi} _{j}^{k}$. 
In order to reduce the high computational cost, by using the first-order approximation
 of Taylor approximation, an approximation of (8) satisfies a linear relation 
 between  $\boldsymbol{\xi} _{ij}^{k}$ and $\Delta \boldsymbol{\xi} _{i}^{k}$, which is given by
\begin{equation}
\label{deqn_ex9a}
\boldsymbol{\xi} _{ij}^{k}\approx \boldsymbol{\xi} _{ij}^{k-1}-\Delta \boldsymbol{\xi} _{i}^{k}+\Delta \boldsymbol{\xi} _{j}^{k}
\end{equation}

The derivation of (9) is given as follows: 

{\bf{Proof 1}} {\textit{Based on the assumption of small perturbation, for}} $\boldsymbol{\xi} \in {{\mathbb{R}}^{6}}${\textit{, the following approximations hold}} \cite{ref27}{\textit{,}}
\begin{equation}
\label{deqn_ex10a}
\log {{\left( \exp \left( \Delta {{\boldsymbol{\xi} }^{\wedge }} \right)\text{exp}\left( {{\boldsymbol{\xi} }^{\wedge }} \right) \right)}^{\vee }}\approx \boldsymbol{\xi} +\boldsymbol{\mathcal{J}}_{l}^{-1}\left( \boldsymbol{\xi} \right)\Delta {{\boldsymbol{\xi} }^{\wedge }}
\end{equation}
\begin{equation}
\label{deqn_ex11a}
\log {{\left( \text{exp}\left( {{\boldsymbol{\xi}  }^{\wedge }} \right)\exp \left( \Delta {{\boldsymbol{\xi} }^{\wedge }} \right) \right)}^{\vee }}\approx\boldsymbol{\xi}  +\boldsymbol{\mathcal{J}}_{r}^{-1}\left(\boldsymbol{\xi}  \right)\Delta {{\boldsymbol{\xi}  }^{\wedge }}
\end{equation}
{\textit{where $\boldsymbol{\mathcal{J}}_{l}^{{}}\left(\boldsymbol{\xi}   \right)\in {{\mathbb{R}}^{6\times 6}}$, $\boldsymbol{\mathcal{J}}_{r}^{{}}\left( \boldsymbol{\xi}   \right)\in {{\mathbb{R}}^{6\times 6}}$ denote the left Jacobian and the right Jacobian of $\boldsymbol{\xi}$, 
respectively. The expression of $\boldsymbol{\mathcal{J}}_{l}^{{}}\left(\boldsymbol{\xi}   \right)$ and $\boldsymbol{\mathcal{J}}_{r}^{{}}\left(\boldsymbol{\xi}   \right)$ can be found in}} \cite{ref28}.

{\textit{Based on the assumption of small perturbation, the first-order Taylor approximation of (8) can be expressed in the form}}
\begin{align}
\label{deqn_ex12a}
\boldsymbol{\xi} _{ij}^{k} {} & =  \boldsymbol{\xi} _{ij}^{k}\left( -\Delta \boldsymbol{\xi} _{i}^{k},\Delta \boldsymbol{\xi} _{j}^{k} \right) \notag \\ 
                  {} &\approx \boldsymbol{\xi} _{ij}^{k}\left( \mathbf{0},\mathbf{0} \right)+\frac{\partial \boldsymbol{\xi} _{ij}^{k}}{\partial \left( -\Delta \boldsymbol{\xi}_{i}^{k} \right)}\left( -\Delta \boldsymbol{\xi} _{i}^{k}-\mathbf{0} \right) \notag  \\
                  & +\frac{\partial \boldsymbol{\xi} _{ij}^{k}}{\partial \Delta \boldsymbol{\xi} _{j}^{k}}\left( \Delta \boldsymbol{\xi} _{j}^{k}-\mathbf{0} \right) \notag \\ 
                  {} & =\boldsymbol{\xi} _{ij}^{k-1}- \boldsymbol{\mathcal{J}}_{l}^{-1}\left( \boldsymbol{\xi}_{ij}^{k-1} \right)\Delta \boldsymbol{\xi} _{i}^{k}+\boldsymbol{\mathcal{J}}_{r}^{-1}\left( \boldsymbol{\xi} _{ij}^{k-1} \right)\Delta\boldsymbol{\xi} _{j}^{k}
\end{align}
{\textit{Notice that $\boldsymbol{\mathcal{J}}_{l}^{-1}\left( \boldsymbol{\xi}_{ij}^{k-1} \right)$ and $\boldsymbol{\mathcal{J}}_{r}^{-1}\left( \boldsymbol{\xi}_{ij}^{k-1} \right)$ are both nonlinear functions of $\boldsymbol{\xi}_{ij}^{k-1}$, changing with the iterations, which also yields a rather high numerical complexity.}}
{\textit{In view of (6) and (7), when the difference between the residual $\Delta \mathbf{M}_{ij}^{k-1}$ and identity matrix $\mathbf{I}\in {{\mathbb{R}}^{4\times 4}}$ is small, we have $\boldsymbol{\xi} _{ij}^{k-1}\approx \mathbf{0}$. It allows us to neglect first and higher order terms of $\boldsymbol{\mathcal{J}}_{l}^{-1}\left( \boldsymbol{\xi}_{ij}^{k-1} \right)$ 
and $\boldsymbol{\mathcal{J}}_{r}^{-1}\left( \boldsymbol{\xi}_{ij}^{k-1} \right)$ to improve the computational efficiency. Under the foregoing conditions, $\boldsymbol{\mathcal{J}}_{l}^{-1}\left( \boldsymbol{\xi}_{ij}^{k-1} \right)$ 
and $\boldsymbol{\mathcal{J}}_{r}^{-1}\left( \boldsymbol{\xi}_{ij}^{k-1} \right)$ can be evaluated up to their approximation, i.e., }} 
\begin{equation}
\label{deqn_ex13a}
\boldsymbol{\mathcal{J}}_{l}^{-1}\left( \boldsymbol{\xi}_{ij}^{k-1} \right)\approx \mathbf{I},\boldsymbol{\mathcal{J}}_{r}^{-1}\left( \boldsymbol{\xi}_{ij}^{k-1} \right)\approx \mathbf{I}
\end{equation}

{\textit{Thus, (9) is derived upon substituting (13) into (12).}}

Note that the term of the left-hand side of (9), i.e., $\boldsymbol{\xi} _{ij}^{k}$, is close to $\mathbf{0}$ when $\Delta \mathbf{M}_{ij}^{k}\approx \mathbf{I}$, 
 henceforth, we can rewrite the equation with the unknowns $\Delta \boldsymbol{\xi} _{i}^{k}$ and $\Delta \boldsymbol{\xi} _{j}^{k}$ , on the left-hand side, as
\begin{equation}
\label{deqn_ex14a}
\boldsymbol{\xi}  _{ij}^{k-1}=\Delta \boldsymbol{\xi}  _{i}^{k}-\Delta \boldsymbol{\xi}  _{j}^{k}
\end{equation}

Specifically, a strategy for improving the robustness of the MA algorithm is to introduce the weight ${{w}_{ij}}$ for each edge $\left( i,j \right)$, 
as adopted in \cite{ref17}. Thus, (14) can be extended as
\begin{equation}
\label{deqn_ex15a}
{{w}_{ij}}\boldsymbol{\xi}  _{ij}^{k-1}={{w}_{ij}}\left(\Delta \boldsymbol{\xi}  _{i}^{k}-\Delta \boldsymbol{\xi}  _{j}^{k}\right)
\end{equation}
Furthermore, let $\Delta \boldsymbol{\xi} _{G}^{k}={{\left[ \begin{matrix}{{\left( \Delta \boldsymbol{\xi} _{1}^{k} \right)}^{\text{T}}} &,&\cdots &,& {{\left( \Delta \boldsymbol{\xi} _{n}^{k} \right)}^{\text{T}}}  \\
\end{matrix} \right]}^{\text{T}}}$. It is the set of Lie algebras corresponding to the increment of global motions. 
$\Delta \boldsymbol{\xi}_{G}^{k}$ contains $6n$ unknowns needed to be determined. In addition, by virtue of $\Delta \boldsymbol{\xi}_{G}^{k}$, (15) can be rewritten in the form
\begin{equation}
\label{deqn_ex16a}
{{w}_{ij}}\boldsymbol{\xi}  _{ij}^{k-1}={\boldsymbol{J}_{ij}}\Delta \boldsymbol{\xi}_{G}^{k}
\end{equation}
where ${\boldsymbol{J}_{ij}}=\left[ \mathbf{0},\cdots ,-w_{ij}^{k}\mathbf{I},\cdots ,w_{ij}^{k}\mathbf{I},\cdots ,\mathbf{0} \right]$ is constructed with the $i$-th and $j$-th block elements filling with 
$-w_{ij}^{k}\mathbf{I}$ and $w_{ij}^{k}\mathbf{I}$. Furthermore, in the absence of outliers, (16) is always satisfied for each edge
$\left(i,j\right) \in E$. Thus, considering all edges in $E$, we can obtain
\begin{equation}
\label{deqn_ex17a}
\boldsymbol{\xi}^{k-1}=\boldsymbol{J}\Delta \boldsymbol{\xi}_{G}^{k}
\end{equation}
where $\boldsymbol{J}\text{=}\left[ \begin{matrix} {\boldsymbol{J}_{ij,1}}; & \cdots ; & {\boldsymbol{J}_{ij,m}} \end{matrix} \right]$ is a $6m\times6n$ matrix concatenated with all matrices $\left\{ {\boldsymbol{J}_{ij,h}} \right\}_{h=1}^{m}$;
$\boldsymbol{\xi} _{{}}^{k-1}=\left[ \begin{matrix} w_{ij,1}^{k}\boldsymbol{\xi} _{ij,1}^{k-1}; & \cdots ; & w_{ij,m}^{k}\boldsymbol{\xi} _{ij,m}^{k-1} \end{matrix} \right]$ 
is the $6m\times1$ column vector obtained by stacking all column vectors $\left\{ w_{ij,h}^{k}\boldsymbol{\xi} _{ij,h}^{k-1} \right\}_{h=1}^{m}$.
Equipped with (17) at hand, most existing MA-based methods, such as MA, WMA and MCC-MA, solve for $\Delta \boldsymbol{\xi}  _{G}^{k}$ as
\begin{equation}
\label{deqn_ex18a}
\Delta \boldsymbol{\xi} _{G}^{k}={{\boldsymbol{J}}^{\dagger }}{{\boldsymbol{\xi} }^{k-1}}
\end{equation}
where ${\boldsymbol{J}}^{\dagger }$ is the pseudo-inverse matrix of ${\boldsymbol{J}}$. Finally, each element $\Delta \boldsymbol{\xi} _{i}^{k}$ of $\Delta \boldsymbol{\xi} _{G}^{k}$ can be utilized to correct the corresponding global motion, i.e., 
\begin{equation}
\label{deqn_ex19a}
\mathbf{M}_{i}^{k}=\mathbf{M}_{i}^{k-1}\exp {{\left( \Delta \boldsymbol{\xi} _{i}^{k} \right)}^{\wedge }}
\end{equation}

In addition, the convergence criterion of WMA or MA can be given by $\left\| \Delta \boldsymbol{\xi}  _{G}^{k} \right\|\le \epsilon$, where $\epsilon$ is a prescribed tolerance.

Although the MA-based algorithms are favored in multi-view registration, they touch upon some common and underlying issues: 

- {\textit{Low efficiency}}. For MA, the matrix $\boldsymbol{J}$ is fixed and does not change with the iteration, and thus, its pseudo-inverse matrix 
${{\boldsymbol{J}}^{\dagger }}$ can be computed before the beginning of the iteration. However, for WMA, if the weight $w_{ij}^{k}$ which denotes the reliability of the relative motion
$\mathbf{M}_{ij}$ is the function of the current residual, and thus WMA needs to update the weights and further update the matrix $\boldsymbol{J}$, which indicates that the WMA algorithm 
needs to compute the pseudo-inverse matrix of $\boldsymbol{J}$ in each iteration. This step leads to high complexity and low efficiency. An 
approach can be used to improve the efficiency of WMA, that is, fixed the weights in advance, such as the operation introduced in 
WMAA \cite{ref17}. An obvious truth is that determining the weights which can accurately reflect the reliability of relative 
motions, is very difficult or even impossible. In addition, the approach does not ensure the performance of WMAA in terms of 
accuracy, the reason for that will be presented in Section IV. 

- {\textit{Local minima and high sensibility to initialization}}. In practice, the number of relative motions is larger than that 
of global motions, and thus, we cannot find an estimate of global motions that exactly satisfies all the constraints given in 
(15), indicating that (17) is an overdetermined system of linear equations. Apparently, for an overdetermined system $\boldsymbol{\xi}^{k-1}=\boldsymbol{J}\Delta \boldsymbol{\xi}_{G}^{k}$, 
$\Delta \boldsymbol{\xi} _{G}^{k}={{\boldsymbol{J}}^{\dagger }}{{\boldsymbol{\xi} }^{k-1}}$ is the unique least-square solution with an orthogonalization procedure. 
In other words, most existing MA-based methods aim at minimizing the following objective function: 
\begin{equation}
\label{deqn_ex20a}
\Delta\boldsymbol{\xi} _{G}^{k}=\underset{\Delta \boldsymbol{\xi} _{G}^{k}}{\mathop{\arg \min }}\,{{\left\| \boldsymbol{\xi}_{{}}^{k-1}-\boldsymbol{J}\Delta \boldsymbol{\xi} _{G}^{k} \right\|}}
\end{equation}

Obviously, these MA-based algorithms solve for the increment $\Delta\boldsymbol{\xi} _{G}^{k}$ based on the 
Gauss-Newton method. It is widely known that the Gauss-Newton algorithm may converge slowly or not all if the initialization is 
far from the minima and be easy to trap in local minima if the initialization is close enough to them. Thus, an initial guess 
 close to ground truth is essential for these MA-based algorithms, indicating that they have 
a high sensibility to initialization. Moreover, the iterative procedure of Gauss-Newton method 
will be stopped when the matrix $\boldsymbol{J}$ is close to being ill-conditioning, regardless of whether the algorithms have converged. In 
addition, the singularity of $\boldsymbol{J}$ may not only cause a convergence problem, but also the loss of accuracy because 
ill-conditioning is prone to numerical instabilities. Both problems pose a reliability or stability problem and even lead to the 
failures of motion averaging. 

\section{Robust Motion Averaging with Laplacian Kernel-Based MCC}
In this section, we briefly introduce LMCC, and propose a robust and efficient motion averaging framework.
 A robust strategy for determining the kernel width of the Laplacian 
function adopted here is also provided.
\subsection{Laplacian Kernel-based Maximum Correntropy Criterion}
Given two random variables $\boldsymbol{X}$ and $\boldsymbol{Y}$, whose joint probability distribution function (PDF) is ${{F}_{\boldsymbol{XY}}}\left( x,y \right)$, the correntropy is defined by \cite{ref21}:
\begin{equation}
\label{deqn_ex21a}
V\left( \boldsymbol{X},\boldsymbol{Y} \right)=E\left[ \kappa \left( \boldsymbol{X},\boldsymbol{Y} \right) \right]=\int{\kappa \left( \boldsymbol{X},\boldsymbol{Y} \right)d}{{F}_{\boldsymbol{X}\boldsymbol{Y}}}\left( x,y \right)
\end{equation}
where $\kappa \left(  \boldsymbol{X},\boldsymbol{Y} \right)$ denotes a shift-invariant Mercer kernel and $E\left[ \kappa \left( \boldsymbol{X},\boldsymbol{Y} \right) \right]$ denotes the 
expectation operator. In practice, the joint PDF is usually unknown, and thus, the conrrentropy can be approximatively estimated through a finite number of data samples 
$\left\{ {{x}_{i}},{{y}_{i}} \right\}_{i=1}^{N}$, that is,
\begin{equation}
\label{deqn_ex22a}
\hat{V}\left( \boldsymbol{X},\boldsymbol{Y} \right)=\frac{1}{N}\sum\limits_{i=1}^{N}{\kappa \left( {{x}_{i}},{{y}_{i}} \right)}
\end{equation}

Instead of the Gaussian kernel adopted in \cite{ref8}, we choose the Laplacian kernel as the kernel, as suggested in \cite{ref24}, which is defined by
\begin{equation}
\label{deqn_ex23a}
\kappa \left( {{x}_{i}},{{y}_{i}} \right)={{\mathcal{L}}_{\sigma }}\left( {{x}_{i}},{{y}_{i}} \right)={{e}^{-\frac{\left| {{x}_{i}}-{{y}_{i}} \right|}{\sigma }}}
\end{equation}
where $\sigma >0$ is the kernel width which controls the radial range of the function, and ${{e}_{i}}={{x}_{i}}-{{y}_{i}}$ is the error.
Moreover, under LMCC, the optimal solution $\boldsymbol{W}$ is found by maximizing the following utility function.
\begin{equation}
\label{deqn_ex24a}
\boldsymbol{W}=\underset{\boldsymbol{W}}{\mathop{\arg \max }}\,\frac{1}{N}\sum\limits_{i=1}^{N}{{{\mathcal{L}}_{\sigma }}\left( {{e}_{i}} \right)}
\end{equation}
\subsection{Motion Averaging under LMCC}
In this subsection, following the derivations and the notations given in Section II, we will solve the WMA problem in multi-view registration based LMCC.

In view of (9), for each scans pair in the given set of relative motions, the following relation holds,
\begin{equation}
\label{deqn_ex25a}
\boldsymbol{\xi} _{ij}^{k}\approx \boldsymbol{\xi} _{ij}^{k-1}+{\boldsymbol{J}_{ij}}\Delta \boldsymbol{\xi} _{G}^{k}
\end{equation}
where $\boldsymbol{\xi} _{ij}^{k}$ can be considered as the errors at $k$-th iteration and ${\boldsymbol{J}_{ij}}=\left[ \mathbf{0},\cdots ,-\mathbf{I},\cdots ,\mathbf{I},\cdots ,\mathbf{0} \right]$ 
constructed with the $i$-th and $j$-th block elements filling with $-\mathbf{I}$ and $\mathbf{I}$. Therefore, upon substitution (25) into (24), the main task at hand, i.e., determining the 
increment of global motions $\Delta \boldsymbol{\xi} _{G}^{k}$, can then be formulated as an optimization problem aimed at maximizing the following object function:
\begin{equation}
\label{deqn_ex26a}
\underset{\Delta \boldsymbol{\xi} _{G}^{k}}{\mathop{\arg \max }}\,\sum\limits_{\left( i,j \right)\in E}{{{\mathcal{L}}_{\sigma }}\left(\boldsymbol{\xi} _{ij}^{k-1}+{\boldsymbol{J}_{ij}}\Delta\boldsymbol{\xi} _{G}^{k} \right)}
\end{equation}
namely,
\begin{equation}
\label{deqn_ex27a}
\underset{\Delta \boldsymbol{\xi}  _{G}^{k}}{\mathop{\arg \max }}\,\sum\limits_{\left( i,j \right)\in E}{{{e}^{-\frac{\left\| \boldsymbol{\xi}  _{ij}^{k-1}+{\boldsymbol{J}_{ij}}\Delta \boldsymbol{\xi}  _{G}^{k} \right\|}{\sigma }}}}
\end{equation}

It is apparent that the objective functions given in (27) and (20) share the same characteristic, namely, considering all 
constraints as a whole, which is also the basic idea of MA algorithm. Furthermore, a simple way of solving the 
optimization problem is by introducing the Half-quadratic optimization technique and dividing it into two subproblems 
to handle \cite{ref24}, as described below:

Define $f\left( y \right)={{e}^{-y}}$, there is a convex conjugate function $\phi$ corresponding to $f\left( y \right)$, and the following relation holds \cite{ref24},
\begin{equation}
\label{deqn_ex28a}
f\left( y \right)=\underset{w}{\mathop{\max }}\,\left( wy-\phi \left( w \right) \right)
\end{equation}
and for a fixed $y$, the maximum is achieved at $w=-f\left( y \right)$.

Thus, from the foregoing relations, if we further define
\begin{equation}
\label{deqn_ex29a}
y_{ij}^{k}=\frac{\left\| \boldsymbol{\xi} _{ij}^{k-1}+{\boldsymbol{J}_{ij}}\Delta \boldsymbol{\xi} _{G}^{k} \right\|}{\sigma }
\end{equation}
we have
\begin{equation}
\label{deqn_ex30a}
f\left( y_{ij}^{k} \right)=\underset{w_{ij}^{k}}{\mathop{\max }}\,\left( w_{ij}^{k}\frac{\left\| \boldsymbol{\xi} _{ij}^{k-1}+{\boldsymbol{J}_{ij}}\Delta \boldsymbol{\xi} _{G}^{k} \right\|}{\sigma }-\phi \left( w_{ij}^{k} \right) \right)
\end{equation}

Now, upon substitution of (30) into (27), an alternative expression of the correntropy measure based 
objective function is derived, namely, 
\begin{equation}
\label{deqn_ex31a}
\begin{aligned}
&  {{J}_{{{\mathcal{L}}_{\sigma }}}}\left( \Delta  \boldsymbol{\xi} _{G}^{k} \right) = \\
&\underset{w_{ij}^{k}}{\mathop{\max }}\,\sum\limits_{\left( i,j \right)\in E}{\left( w_{ij}^{k}\frac{\left\|  \boldsymbol{\xi} _{ij}^{k-1}+{ \boldsymbol{J}_{ij}}\Delta  \boldsymbol{\xi}_{G}^{k} \right\|}{\sigma }-\phi \left( w_{ij}^{k} \right) \right)}
\end{aligned}
\end{equation}
Further, we can define the augmented cost function:
\begin{equation}
\label{deqn_ex32a}
\begin{aligned}
& {{J}_{HQ}}\left( \Delta \boldsymbol{\xi} _{G}^{k},{\boldsymbol{w}^{k}} \right)=  \\
& \sum\limits_{\left( i,j \right)\in E}{\left( w_{ij}^{k}\frac{\left\| \boldsymbol{\xi} _{ij}^{k-1}+{\boldsymbol{J}_{ij}}\Delta \boldsymbol{\xi}_{G}^{k} \right\|}{\sigma }-\phi \left( w_{ij}^{k} \right) \right)}
\end{aligned}
\end{equation}
where ${\boldsymbol{w}^{k}}=\left[ \begin{matrix}  w_{ij,1}^{k}; & \cdots  & w_{ij,m}^{k} \end{matrix} \right]$ is a $m\times1$ vector.

According to the foregoing discussion, we can obtain the equivalent relation as follows:
\begin{equation}
\label{deqn_ex33a}
\underset{\Delta \boldsymbol{\xi} _{G}^{{k}}}{\mathop{\max }}\,{{J}_{{{\mathcal{L}}_{\sigma }}}}\left( \Delta \boldsymbol{\xi} _{G}^{k} \right)=\underset{\Delta \boldsymbol{\xi}_{G}^{k},{\boldsymbol{w}^{k}}}{\mathop{\max }}\,{{J}_{HQ}}\left( \Delta \xi _{G}^{k},{\boldsymbol{w}^{k}} \right)
\end{equation}

Hence, we can determine the increment $\Delta \boldsymbol{\xi} _{G}^{k}$ to solve the following maximization problem:
\begin{equation}
\label{deqn_ex34a}
\begin{aligned}
&\Delta \boldsymbol{\xi} _{G}^{k} =\underset{\Delta \boldsymbol{\xi} _{G}^{k},{\boldsymbol{w}^{k}}}{\mathop{\arg \max }}\,{{J}_{HQ}}\left( \Delta \boldsymbol{\xi} _{G}^{k},{\boldsymbol{w}^{k}} \right) \\ 
 & =\underset{\Delta \boldsymbol{\xi} _{G}^{k},{\boldsymbol{w}^{k}}}{\mathop{\arg \max }}\,\sum\limits_{\left( i,j \right)\in E}{\left( w_{ij}^{k}\frac{\left\| \boldsymbol{\xi} _{ij}^{k-1}+{\boldsymbol{J}_{ij}}\Delta \boldsymbol{\xi} _{G}^{k} \right\|}{\sigma }-\phi \left( w_{ij}^{k} \right) \right)}  
\end{aligned}
\end{equation}
Treating $\Delta \boldsymbol{\xi} _{G}^{k}$ and $\boldsymbol{w}^{k}$ as a whole, the optimization problem can readily be solved by using the method proposed in \cite{ref24}. 
All these steps of the optimization method are given below:

1) optimization of $\boldsymbol{w}^{k}$: According to (28) and (29), once we can observe a certain $\bar{y}_{ij}^{k-1}={}^{\left\| \boldsymbol{\xi} _{ij}^{k-1} \right\|}/{}_{{{\sigma }^{k}}}$, we can obtain a good approximation to the solution of (28) which is 
$w_{ij}^{k}=-{{e}^{-\bar{y}_{ij}^{k-1}}}$. Therefore, considering each edge in $E$, the optimal solution $\boldsymbol{w}^{k}$ of (34) can be determined as
\begin{equation}
\label{deqn_ex35a}
{{\bar{\boldsymbol{w}}}^{k}}=\left\{ \left. \bar{{w}}_{ij}^{k}=-{{e}^{-\bar{y}_{ij}^{k-1}}} \right|\forall \left( i,j \right)\in E \right\}
\end{equation}

Apparently, the significance of (35) is that each relative motion can be assigned a different weight in the current iteration automatically based on the corresponding residual
$\boldsymbol{\xi} _{ij}^{k-1}$ according to the property of the Laplacian function. In addition, the weight assignment operation also implies the close relationship between 
correntropy measure and robust M-estimation, which allows us to eliminate the effect caused by outliers and enhance the robustness of WMA algorithm. It is also noteworthy 
that the reliability of $\bar{w}_{ij}^{k}$ depends on the accuracy of the current residual $\boldsymbol{\xi} _{ij}^{k-1}$ for estimation of the noise level of pair-wise 
registration results. Moreover, when the initial motions $\left\{ \mathbf{M}_{i}^{0} \right\}_{i=1}^{n}$ is close to the ground truth $\left\{ \mathbf{M}_{i}^{*} \right\}_{i=1}^{n}$,  
the residual $\boldsymbol{\xi} _{ij}^{k-1}$  can accurately reflect the noise level of the registration result of the scans pair $\left( i,j \right)$, and hence, the weight 
$\bar{w}_{ij}^{k}$ has rather high reliability. However, once $\left\{ \mathbf{M}_{i}^{0} \right\}_{i=1}^{n}$ is far from $\left\{ \mathbf{M}_{i}^{*} \right\}_{i=1}^{n}$, the residual $\boldsymbol{\xi } _{ij}^{0}$ 
will be wrongly determined for each pair-wise result at the first iteration and $\bar{w}_{ij}^{k}$ is also wrongly estimated, which even makes the failure of motion averaging. As a consequence, based on the foregoing perception, 
the properties of correntropy measure also imply the dependence on a good initial motions. In other words, our method is also a local convergent algorithm. 

2) Optimization of $\Delta \boldsymbol{\xi} _{G}^{k}$: Given fixed ${{\bar{\boldsymbol{w}}}^{k}}$, (34) is simplified into the following optimization problem: 
\begin{equation}
\label{deqn_ex36a}
\Delta \boldsymbol{\xi} _{G}^{k}=\underset{\Delta \boldsymbol{\xi} _{G}^{k}}{\mathop{\arg \max }}\,\sum\limits_{\left( i,j \right)\in E}{\left( \bar{w}_{ij}^{k}\frac{\left\| \boldsymbol{\xi} _{ij}^{k-1}+{\boldsymbol{J}_{ij}}\Delta \boldsymbol{\xi} _{G}^{k} \right\|}{\sigma }-\phi \left( \bar{w}_{ij}^{k} \right) \right)}
\end{equation}
Since $\phi \left( \bar{w}_{ij}^{k} \right)$ is a function of ${{\bar{w}}_{ij}^{k}}$ and not related to $\Delta \boldsymbol{\xi} _{G}^{k}$, the foregoing problem can be further simplified by 
\begin{equation}
  \label{deqn_ex37a}
\Delta \boldsymbol{\xi} _{G}^{k}=\underset{\Delta \boldsymbol{\xi} _{G}^{k}}{\mathop{\arg \max }}\,\sum\limits_{\left( i,j \right)\in E}{\bar{w}_{ij}^{k}\left\| \boldsymbol{\xi} _{ij}^{k-1}+{\boldsymbol{J}_{ij}}\Delta \boldsymbol{\xi} _{G}^{k} \right\|}
\end{equation}

It is easily seen that this problem can be transformed, by introducing an upper bound $\gamma _{ij}^{k}$ for the objective function 
$\left\| \boldsymbol{\xi} _{ij}^{k-1}+{\boldsymbol{J}_{ij}}\Delta \boldsymbol{\xi} _{G}^{k} \right\|$, into an equivalent problem of the following form
\begin{equation}
\label{deqn_ex38a}
\begin{aligned}
  & \underset{\Delta \boldsymbol{\xi} _{G}^{k}}{\mathop{\min }}\,\sum\limits_{\left( i,j \right)\in E}{-{{{\bar{w}}}_{ij}^{k}}\gamma _{ij}^{k}}  \\ 
 & \text{s}\text{.t}\text{. }\left\| \boldsymbol{\xi} _{ij}^{k-1}+{\boldsymbol{J}_{ij}}\Delta \boldsymbol{\xi} _{G}^{k} \right\|\le \gamma _{ij}^{k}   \\ 
\end{aligned}
\end{equation}

It is obvious that, by introducing additional variables, the problem can be efficiently solved using Second Order Cone 
Programming (SOCP) which has global convergence. In conclusion, instead of using Gauss-Newton method for obtaining the
increment of global motions, we obtain that by solving the two subproblems with alternate iterations, which avoids the 
expensive calculation of the pseudo-inverse of a large matrix in the procedure of Gauss-Newton method. Furthermore, in 
views of (38) and (25), for each relative motion, its corresponding ${\boldsymbol{J}_{ij}}$ is fixed, which further reduces the computational cost.

\subsection{Strategy for Determining Kernel Width}
Similar to MCC-MA \cite{ref8}, the performance of the proposed method is affected by the kernel width. Lots of works have 
illustrated that the relatively large kernel width can offer high convergence speed but suffer from less accuracy, and 
vice versa \cite{ref8}. Furthermore, it is important to emphasize that the kernel width must be given appropriately. If the kernel 
width is too small, under the characteristic of the Laplacian kernel function, the weights of most of relative motions 
will be very close to zero, even though these relative motions are relatively “reliable”. This means that the averaged algorithm
does not exploit fully the information redundancy supplied by relative motions. In turn, a large width kernel will result in the invalid exploitation of 
information redundancy because many outliers are considered as inliers. From 
the foregoing discussion, an important conclusion can be drawn that a small kernel leads to the loss of registration results, while a large kernel width reduces 
the robustness to outliers. Thus, an appropriate kernel width is essential for the proposed method to guarantee 
accuracy and robustness. 

In order to balance the accuracy and robustness, a new method to determine the kernel width is introduced 
in this paper, given as follows:
\begin{equation}
\label{deqn_ex39a}
{{\sigma }_{k}}=\max \left( \text{median}\left( {{\left\{ \left\| \boldsymbol{\xi} _{ij}^{k-1} \right\| \right\}}_{\alpha }} \right),\chi  \right)
\end{equation}
where ${{\sigma }_{k}}$ is the kernel width at $k$-th iteration and ${{\left\{ \left\| \boldsymbol{\xi} _{ij}^{k-1} \right\| \right\}}_{\alpha }}$ is 
the set composed of the top ${\alpha }$ percent of elements of ${{\left\{ \left\| \boldsymbol{\xi} _{ij}^{k-1} \right\| \right\}}}$
sorted in ascending order; for a set $\mathbf{Z}$, $\text{median}\left(\mathbf{Z}\right)$ stands for its median value; $\chi $ is constant and 
does not associate with iteration number, which can be considered as the upper bound of ${{\sigma }_{k}}$ to avoid a very small kernel width. 
More specifically, ${\alpha }$ plays a crucial role in judging the reliability of relative motions and utilizing the information redundancy adequately and accurately.

In addition, experimental results show that the performance of our method is not sensitive to the numerical values of $\chi $. More specifically, we found out 
that ${\alpha }$ lies in the range $\left[0.6, 0.7\right]$ is appropriate and suggest that the value of $\chi $ is 0.001. Furthermore, we indeed seen that 
adjusting the value of ${\alpha }$, according to the outlier ratio of relative motions, will improve the performance of our 
method. However, the truth is that the difficulty of knowing the outlier ratio without any prior information does 
not allow us to adjust ${\alpha }$ dynamically. Thus, for simplification we set ${\alpha=0.7}$. Obviously, at the beginning of the iteration, 
the kernel width is relatively large and decreases with the increase of the iteration number, which efficiently balances 
accuracy and efficiency.

Algorithm 1 summarizes the process of the proposed robust motion averaging algorithm for multi-views registration and 
we named the proposed method as L-MA for the sake of simplicity.
\begin{algorithm}[H]
  \renewcommand{\algorithmicrequire}{\textbf{Input:}}  
  \renewcommand{\algorithmicensure}{\textbf{Output:}} 
  \caption{\textbf{L-MA} : robust motion algorithm based LMCC}
  \begin{algorithmic}[1]
  \REQUIRE initial global motions $\left\{ \textbf{M}_{i}^{0} \right\}_{i=1}^{n}$; relative motions $\left\{ {{\textbf{M}}_{ij,h}} \right\}_{h=1}^{m}$;
  \ENSURE an estimate of global motions $\left\{ \textbf{M}_{i}^{{}} \right\}_{i=1}^{n}$;
  \STATE {\text{Initialization:}} $\chi =0.001$, $\alpha =0.7$, $\epsilon  ={{10}^{-4}}$, ${{k}_{\max }}=50$
  \STATE $k\leftarrow 0$
  \STATE \textbf{repeat:}
  \STATE \hspace{0.25cm} $k\leftarrow k+1$
  \STATE \hspace{0.25cm} \textbf{for} $h=1:m$ \textbf{do} 
  \STATE \hspace{0.5cm} Compute the residual error \\ 
  ~~~~~$\boldsymbol{\xi} _{ij}^{k-1}=\log {{\left( {{\left( \mathbf{M}_{i}^{k} \right)}^{-1}}\mathbf{M}_{ij,h}^{{}}\mathbf{M}_{j}^{k} \right)}^{\vee }}$
  \STATE \hspace{0.25cm} \textbf{end} 
  \STATE \hspace{0.25cm} Compute the weights ${\bar{\boldsymbol{w}}}^{k}$ by (34)
  \STATE \hspace{0.25cm} Solve for $\Delta \boldsymbol{\xi} _{G}^{k}$ from the SOCP problem given in (39)
  \STATE \hspace{0.25cm} Correct global motions: $\mathbf{M}_{i}^{k}=\mathbf{M}_{i}^{k-1}\exp  {{\left( \Delta \boldsymbol{\xi} _{i}^{k} \right)}^{\wedge }}$
  \STATE \textbf{until} $k\ge {{k}_{\max }}$ or $\left\| \Delta \boldsymbol{\xi} _{G}^{k} \right\|\le \epsilon $
  \STATE \textbf{return}  $\left\{ \textbf{M}_{i}^{k} \right\}_{i=1}^{n}$
  \end{algorithmic}
  \end{algorithm}
\section{Experiments}
In this section, we evaluate the proposed method on both synthetic and data sets. To verify its performance, we compared our method with four 
related methods mentioned in Section I. The methods are as follows: -the original motion averaging algorithm \cite{ref15} (denoted as MA). The algorithm 
first introduces motion averaging to recover global motions from relative motions but does not consider the reliability of relative motions. 
-the low rank and sparse matrix decomposition method \cite{ref19} (denoted as LRS\footnote[1]{ http://www.diegm.uniud.it/fusiello/demo/gmf}). The method introduces the application of matrix decomposition to 
recover global motions from a set of relative motions for the first time. -the weighted motion averaging algorithm \cite{ref17} (denoted as WMAA\footnote[2]{The source code was kindly provied by the authors of \cite{ref17}}). 
The method introduces the overlapping percentage of each scans pair as the weight of the corresponding relative motion. -the robust 
motion averaging algorithm based maximum correntropy criterion \cite{ref8} (denoted as MCC-MA\footnote[3]{The source code was kindly provied by the authors of \cite{ref8}}). The method introduces the correntropy measure 
to reformulate the MA problem and improves the performance of WMA in terms of accuracy and robustness. All methods are implemented in 
MATLAB and we used the default parameters specified in the source code. In addition, we solved the SOCP problem given in (39) by SeDuMi \cite{ref29}. 
All experiments are performed on a laptop with a 2.90GHZ processor of 16-core and 16GB RAM.

To quantitatively compare the accuracy of different methods, we use the following metrics to evaluate the rotation and translation errors 
of registration results \cite{ref8}:
\begin{equation}
\label{deqn_ex40a}
\begin{aligned}
  & {{e}_{\mathbf{R}}}=\frac{1}{n}\sum\limits_{i=1}^{n}{\text{arcos}}\left( \frac{\operatorname{tr}\left( {{\mathbf{R}}_{i}}\mathbf{R}_{*,i}^{\text{T}} \right)-1}{2} \right) \\ 
 & {{e}_{\mathbf{R}}}=\frac{1}{n}\sum\limits_{i=1}^{n}{{{\left\| {\mathbf{t}_{i}}-{\mathbf{t}_{*,i}} \right\|}}} \\ 
\end{aligned}
\end{equation}
where ${{\mathbf{M}}_{*,j}}=\left\{ {{\mathbf{R}}_{*,i}},{{\mathbf{t}}_{*,i}} \right\}$ and $\left\{ {{\mathbf{R}}_{i}},{{\mathbf{t}}_{i}} \right\}$ denote the ground truth and the estimated global motion 
of $i$-th scan, respectively. In addition, we also 
define the rotation and translation errors of the set of relative motions to quantify the outliers ratio or/and noise 
level, as given by:
\begin{equation}
\label{deqn_ex41a}
\begin{aligned}
  & {{e}_{\mathbf{R},\text{Data}}}=\frac{1}{m}\sum\limits_{i=1}^{m}{\text{arcos}}\left( \frac{\operatorname{tr}\left( {{{\tilde{\mathbf{R}}}}_{ij}} \right)-1}{2} \right) \\ 
 & {{e}_{\mathbf{t},\text{Data}}}=\frac{1}{m}\sum\limits_{i=1}^{m}{\left\| {{{\tilde{\mathbf{t}}}}_{ij}} \right\|} \\ 
\end{aligned}
\end{equation}
where ${{\tilde{\mathbf{R}}}_{ij}}$ and ${\tilde{\mathbf{t}}_{ij}}$ is the rotation matrix and translation vector of ${\tilde{\mathbf{M}}_{ij}}=\mathbf{M}_{*,i}^{-1}{\mathbf{M}_{ij}}{\mathbf{M}_{*,j}}$, 
here, $\mathbf{M}_{ij}$ is the estimated relative motion between $i$-th scan and $j$-th scan and $m$ stands for the total number of scan pairs. 

\subsection{Synthetic data}
To comprehensively compare the performance of all methods, such as robustness to noise and outliers, runtime, sensitivity
to initialization, and so on, all compared methods except for WMAA are tested on synthetic data first because the overlap
percentages of scan pairs, which are necessary for WMAA, cannot be determined under the simulated test. Experimental 
Settings and definitions are described as follows:

1) First, similar to the experimental settings adopted in \cite{ref20}, we generate $n$ global motions $\left\{\mathbf{M}_{i}^{*}=\left(\mathbf{R}_{i}^{*},\mathbf{t}_{i}^{*} \right) \right\}_{i=1}^{n}$ in which the 
rotation matrix $\mathbf{R}_{i}^{*}$ was sampled from random Euler angles, and the elements of the translation vector $\mathbf{t}_{i}^{*}$ followed the standard normal distribution. 
Moreover, since MA, MCC-WMA and our method all require the initial global motions (InitM) as input, noise is added to 
the ground truth to obtain InitM for simplicity. The generation model of InitM is as follows:
\begin{equation}
\label{deqn_ex42a}
\begin{cases}
\mathbf{R}_{i}^{init}=svd\left( \mathbf{R}_{i}^{*}+\sigma _{\mathbf{R}}^{init}{{\boldsymbol{W}}_{\mathbf{R}}} \right) \\
\mathbf{t}_{i}^{init}=\mathbf{t}_{i}^{*}+\sigma _{\mathbf{t}}^{init}{\boldsymbol{W}_{\mathbf{t}}} \\ 
\end{cases} 
\end{equation}
where ${{\boldsymbol{W}}_{\mathbf{R}}}$ and ${\boldsymbol{W}_{\mathbf{t}}}$ are, respectively, a $3\times3$ matrix and a $3\times1$ vector whose elements follow the standard normal 
distribution; $\sigma _{\mathbf{R}}^{init}$ and $\sigma _{\mathbf{t}}^{init}$ are constants standing for the noise levels on rotation and translation of ground 
truth; the operator $svd\left(\mathbf{R} \right)$ denotes the  projection of matrix $\mathbf{R}$ onto $SO(3)$ through SVD so as to render $\mathbf{R}_{i}^{init}$  
orthogonal.

\begin{figure*}[!b]
  \centering
  \subfloat
  {
    \begin{minipage}[t]{0.45\linewidth}
      \centering
      \includegraphics[width=3.25in]{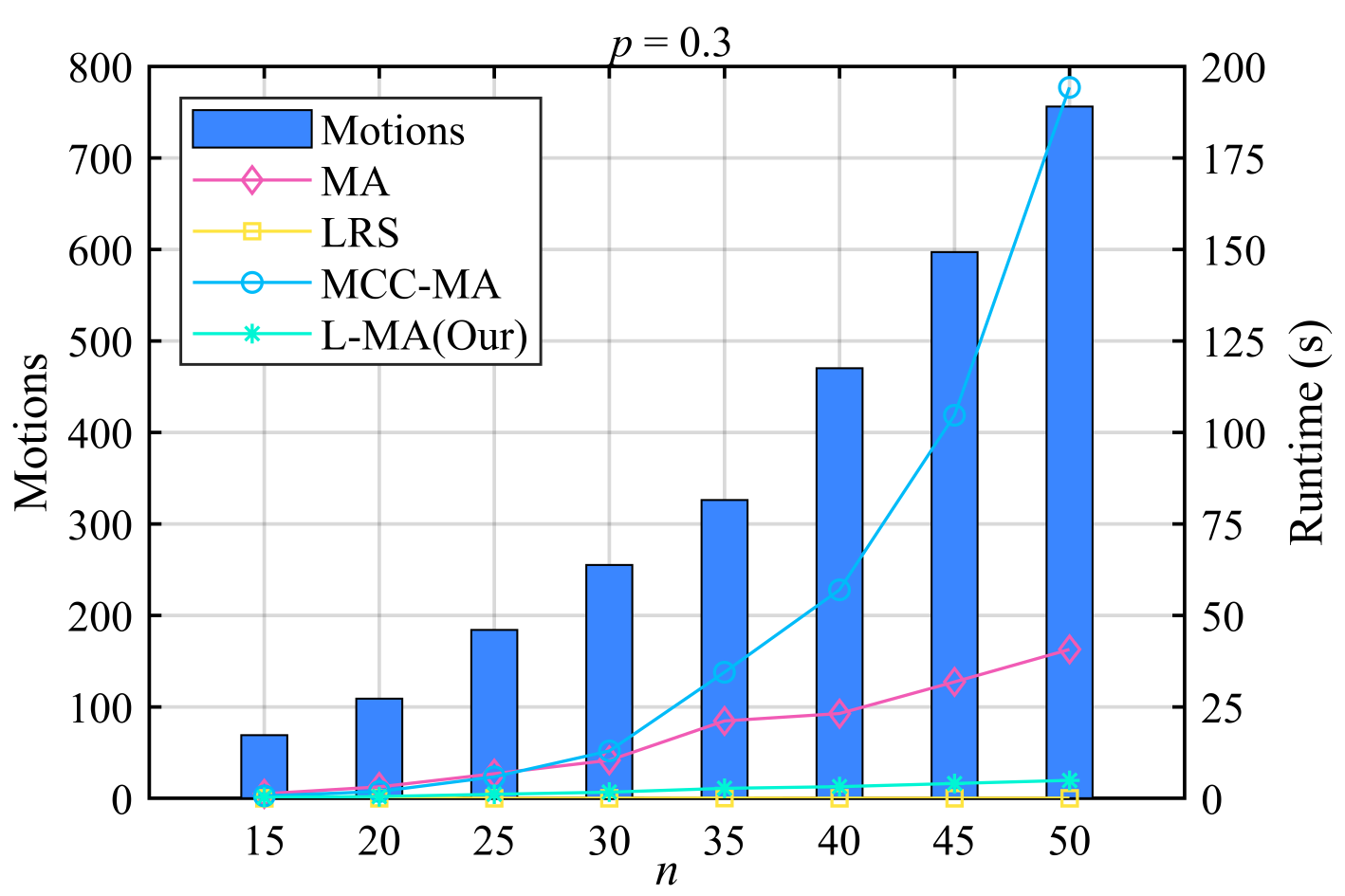} \\
    \end{minipage}
  }
  \subfloat
  {  \begin{minipage}[t]{0.60\linewidth}
      \centering
      \includegraphics[width=3.25in]{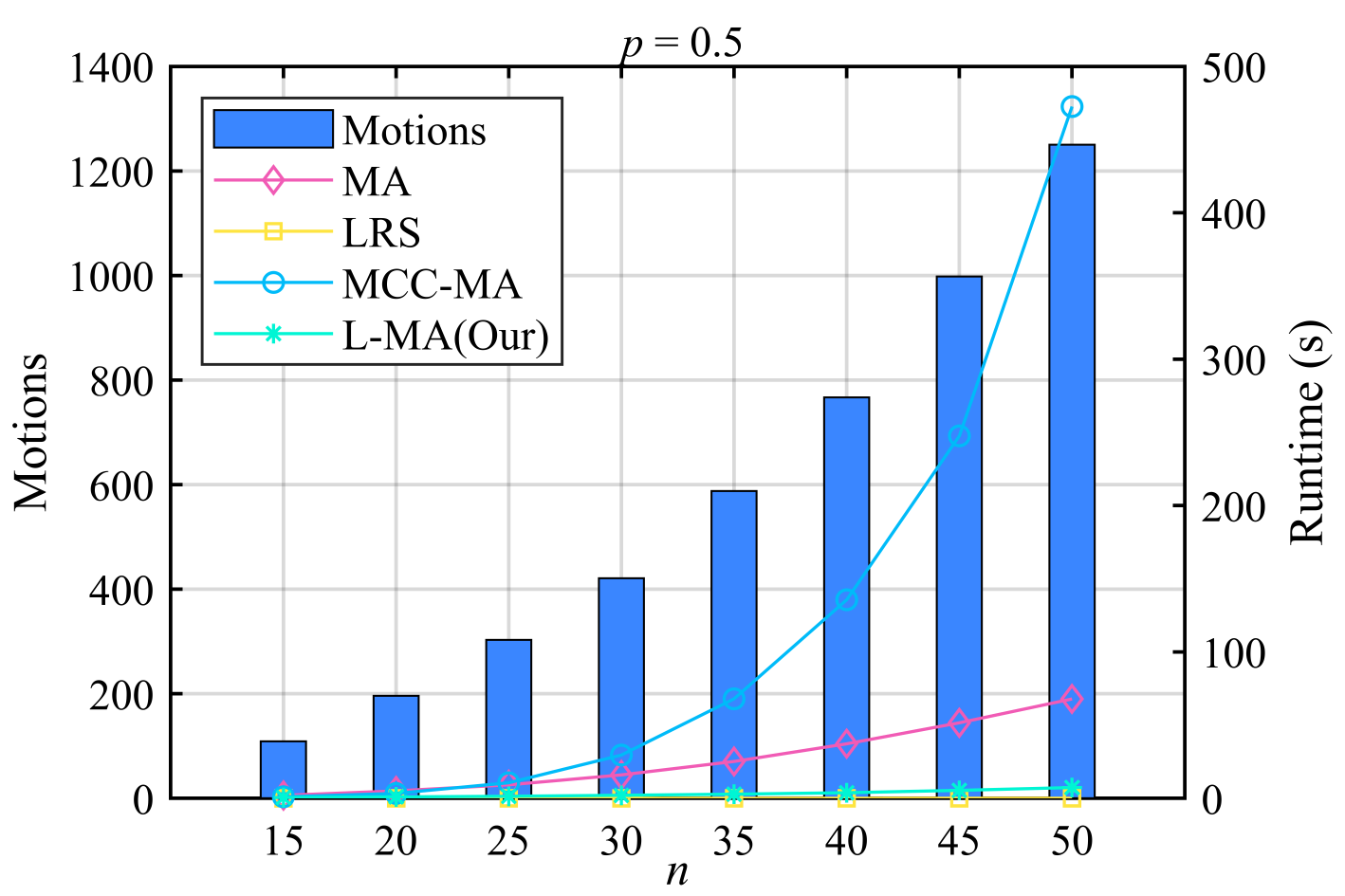} \\
  \end{minipage}
  }
  \caption{Runtimes of different methods (in seconds) and the numbers of relative motions. All results are plotted vs. $n$, with $p=0.3$ (left) and $p=0.5$ (right). }
  \label{fig1}
  \end{figure*}
  \begin{figure*}[!b]
    \centering
    \subfloat
    {
      \begin{minipage}[t]{0.45\linewidth}
        \centering
        \includegraphics[width=3.25in]{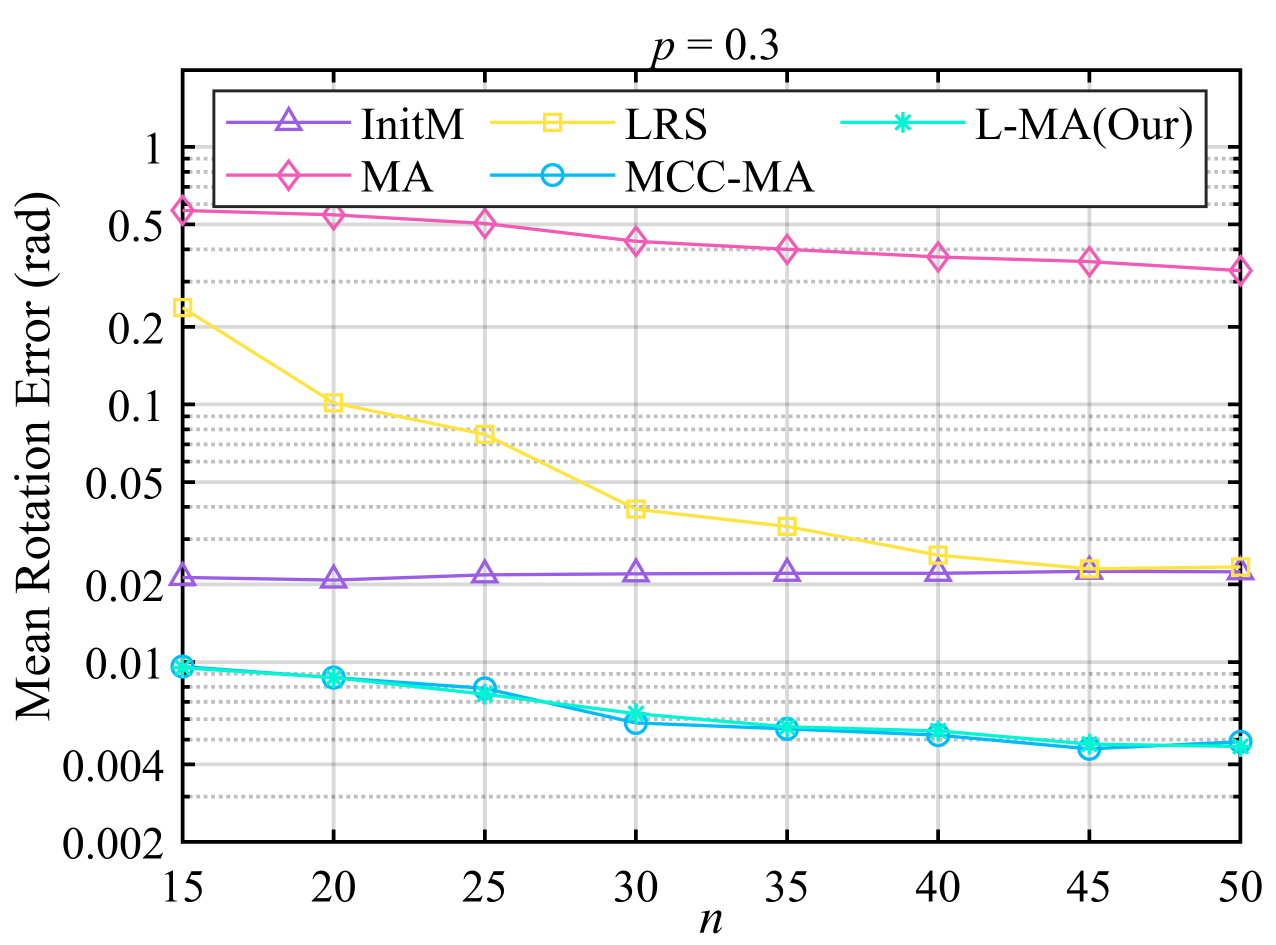} \\
        \includegraphics[width=3.25in]{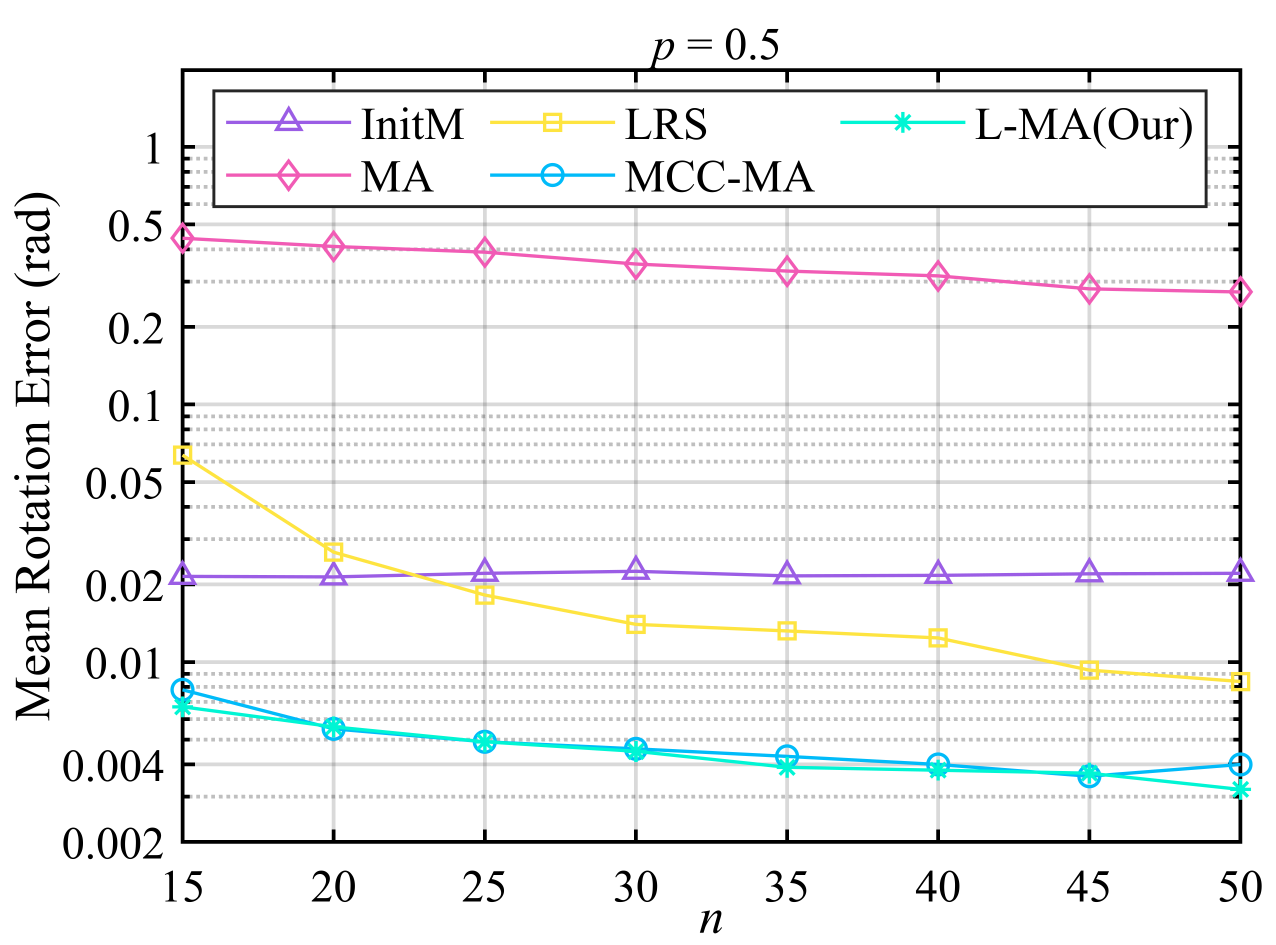}
      \end{minipage}
    }
    \subfloat
    {  \begin{minipage}[t]{0.55\linewidth}
        \centering
        \includegraphics[width=3.25in]{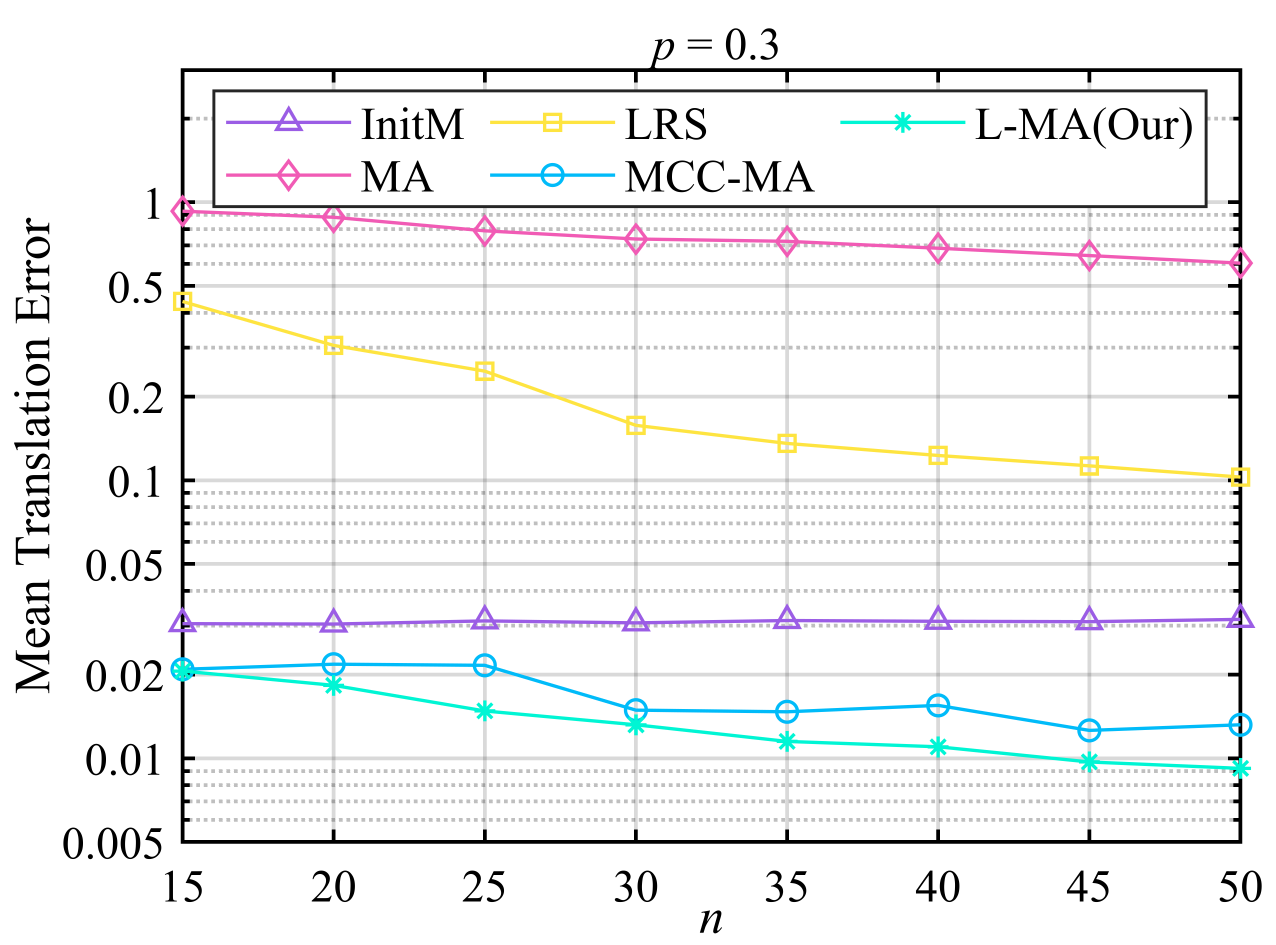} \\
        \includegraphics[width=3.25in]{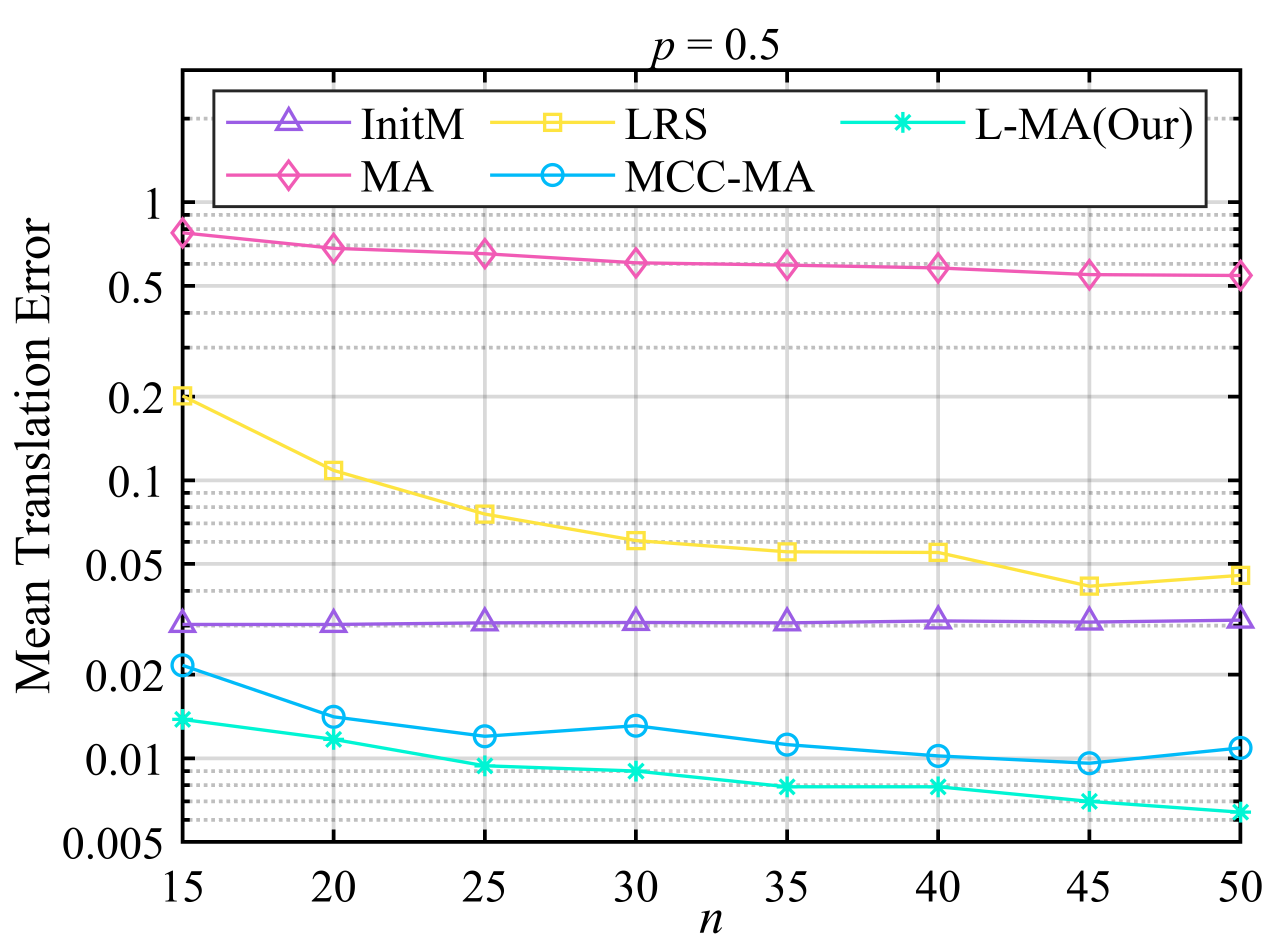}
    \end{minipage}
    }
    \caption{Performance comparison among different methods under different number of views $n$. 
    The mean errors are plotted vs. $n$, with $p=0.3$ (top) and $p=0.5$ (bottom).}
    \label{fig2}
    \end{figure*}

2). As all methods mentioned in this subsection require relative motions as input, we obtained the view-graph of 
relative motions by the Erdős-Rényi model with parameters $\left( n,p \right)$, where $p$ can be considered as the 
probability that there exists edge $\left( i,j \right)$ connecting node $i$ and $j$. Furthermore, for the studies of the 
robustness to noise and outliers of different methods, we generate the outliers from the original relative motions 
with parameter $q$ which denotes the probability that one edge is the outlier. For example, half of the relative 
motions are corrupted for the case in which $q=0.5$.  The generation model of motions of these outliers is similar to that 
of global motions. Moreover, the motions of inliers are generated with parameters $\sigma _{\mathbf{R}}^{inlier}$ and $\sigma _{\mathbf{t}}^{inlier}$ 
which can be expressed as  
\begin{equation}
\label{deqn_ex43a}
\begin{cases}
\mathbf{R}_{ij}^{inlier}=svd\left( \mathbf{R}_{ij}^{*}+\sigma _{\mathbf{R}}^{inlier}{{\boldsymbol{W}}_{\mathbf{R}}} \right) \\
\mathbf{t}_{ij}^{inlier}=\mathbf{t}_{ij}^{*}+\sigma _{\mathbf{t}}^{inlier}{\boldsymbol{W}_{\mathbf{t}}} \\ 
\end{cases} 
\end{equation}
where $\mathbf{R}_{ij}^{*}$ and $\mathbf{t}_{ij}^{*}$ denote, respectively, the rotation matrix and the translation 
vector of the relative motion $\mathbf{M}_{ij}^{*}={{\left( \mathbf{M}_{j}^{*} \right)}^{-1}}\mathbf{M}_{i}^{*}$;  
$\sigma _{\mathbf{R}}^{inlier}$, $\sigma _{\mathbf{t}}^{inlier}$ stand for the noise level of rotation and translation of inliers, respectively.

From the foregoing settings and definitions, the simulated model used in the experiment is affected by parameters 
$\left\{ n,p,q,\sigma _{\mathbf{R}}^{init},\sigma _{\mathbf{t}}^{init},\sigma _{\mathbf{R}}^{inlier},\sigma _{\mathbf{t}}^{inlier} \right\}$.
Obviously, several parameters do not allow us to take all the performance indices into account during the comparative 
experiment. Hence, similar to lots of works of literature, in the following experiments, we changed one part of the 
parameters, while the others remained fixed, to explore the influence of methods by different parameters, so as 
to evaluate the performances of methods. Additionally, in order to eliminate the randomness of simulation data, 
the results reported in the following subsections are the average values of 50 experiments with the same parameters, 
unless otherwise specified.

\subsubsection{Efficiency and Accuracy}
To validate the efficiency and accuracy of our method, we compared the different methods for a set of test parameters. 
In this experiment, we let $n$ vary between 15 and 55, in which case $p=0.3$ and $0.5$. More specifically, the larger $n$ and $p$, 
the greater the number of relative motions needed to be averaged. In both cases, $\sigma _{\mathbf{R}}^{inlier}=\sigma _{\mathbf{t}}^{inlier}=0.01$,
$\sigma _{\mathbf{R}}^{init}=\sigma _{\mathbf{t}}^{init}=0.02$ and $q=0.3$. Under the condition that $p$ is fixed, apparently, the indices, such as 
computation time, and averaged error, can be considered as functions of $n$.

\begin{figure*}[!b]
  \renewcommand{\dblfloatpagefraction}{.9}
\centering
\subfloat
{
  \begin{minipage}{0.45\linewidth}
    \centering
    \includegraphics[width=3.25in]{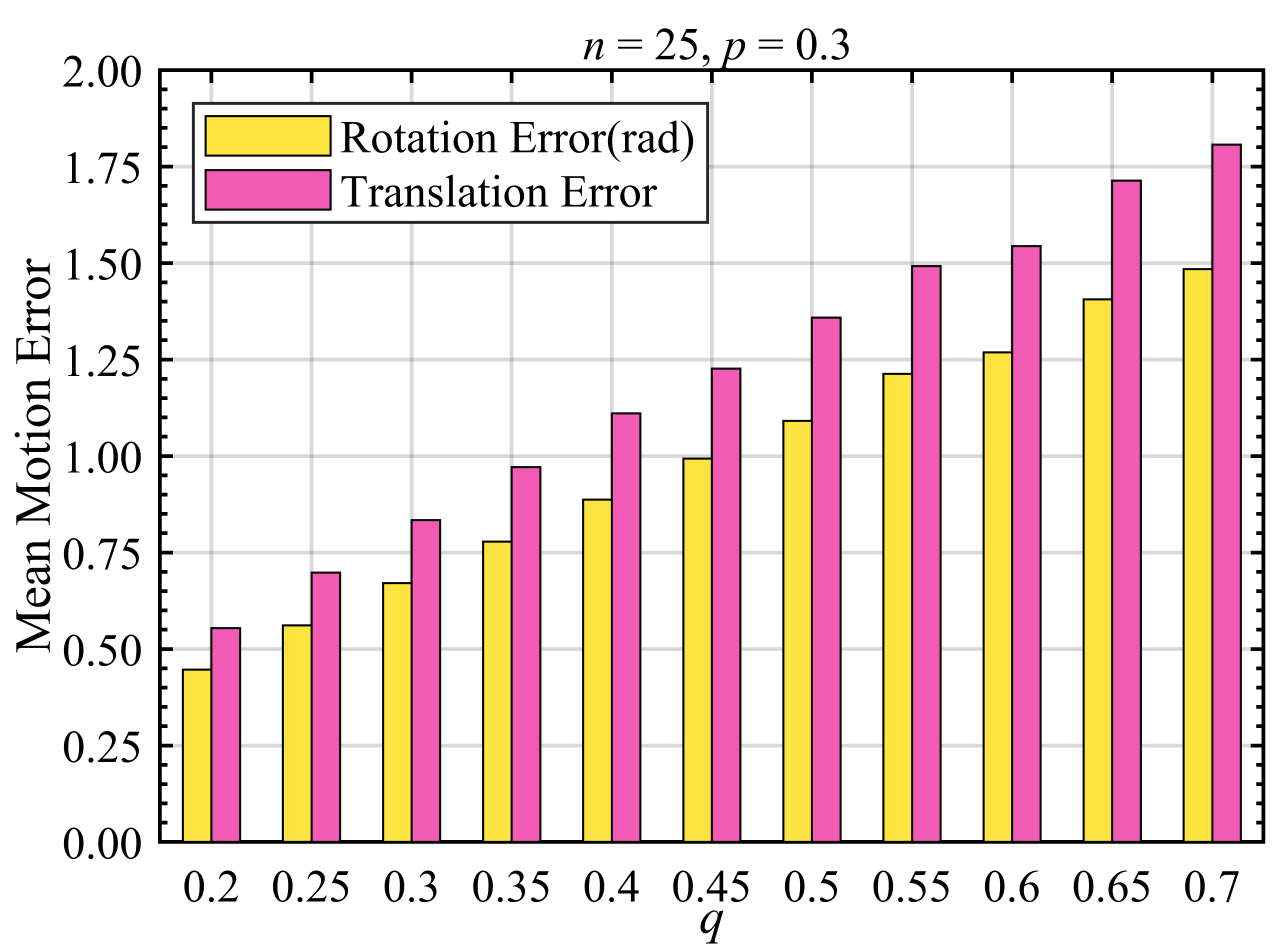} 
  \end{minipage}
}
\subfloat
{  \begin{minipage}{0.55\linewidth}
    \centering
    \includegraphics[width=3.25in]{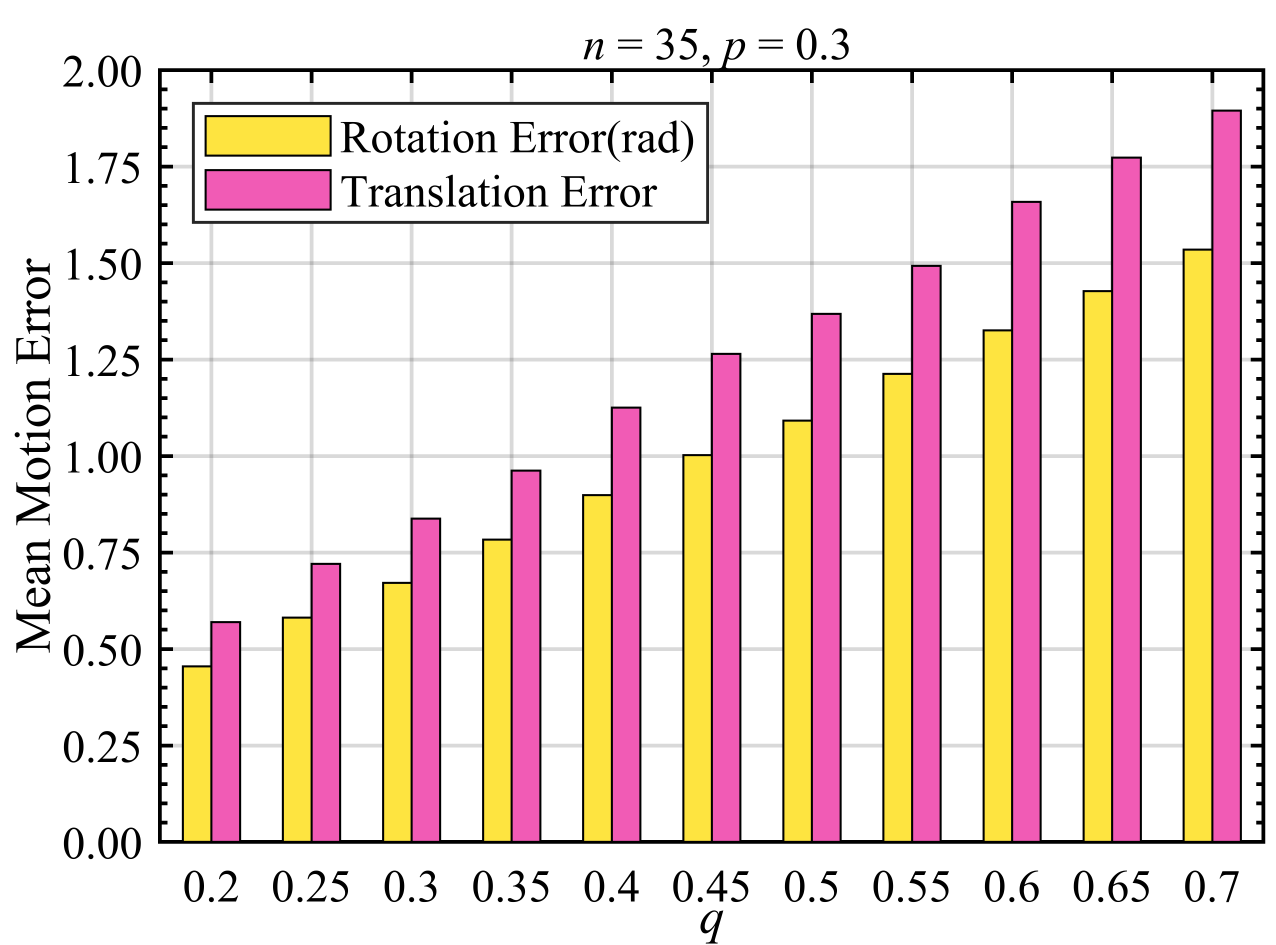} 
\end{minipage}
}
\caption{Mean motion error of relative motions under different outlier ratios. The mean rotation and translation errors are displayed vs. the outlier ratio $q$, with $n=25$ (left) and $n=35$ (right). 
Obviously, as we increase outlier ratio, the motion errors of relative motions increase almost linearly.}
\label{fig3}
\end{figure*}

The averaging runtimes, and numbers of relative motions are depicted in Fig.1. The mean errors of rotation and 
translation of different methods are plotted in Fig.2. As illustrated in Fig.2, our method and 
MCC-MA achieve the lowest rotation error in all simulated data, and our method can always achieve the 
lowest translation error than the other methods. In addition, MA achieves the worst accuracy in 
both rotation and translation. Compared with MA, LRS obtains the more accurate results and achieves the 
third highest accuracy.

In detail, the performance of MA is more affected than other methods by outliers. Therefore, the 
accuracy of results achieved by MA is significantly reduced even when the set of relative motions only contains 
one outlier \cite{ref14}. In addition, LRS is more sensitive to the sparsity of the view-graph than the other methods, this 
is the reason why there is a limit on the accuracy of LRS. Moreover, by virtue of the property of correntropy 
measure, our method and MCC-MA can assign different weight to each relative motion automatically so as to pay 
attention to the reliable relative motions and achieve more accurate results compared with MA and LRS. 

Furthermore, we have also observed in Fig.1 and Fig.2 an interesting phenomenon: the mean errors of all methods 
decrease with the increase of the number of absolute motions $n$ or/and the value of $p$ when the other parameters 
are fixed. The reason for this phenomenon is that the increase of $n$ and/or $p$ will lead to the increase in the 
relevance of each node of the view-graph, that is, the increase in information redundancy, so that the errors of each absolute motion can be better averaged under the 
characteristics of the motion average algorithm. We would like to emphasize that the phenomenon only occurs in the 
simulation case where the outlier ratio and noise levels are fixed. In practice, the outlier ratio and noise levels of relative motions vary with the number of relative motions.

\begin{figure*}[!t]
  \renewcommand{\dblfloatpagefraction}{.9}
    \centering
  \subfloat
  {
    \begin{minipage}{0.45\linewidth}
      \centering
      \includegraphics[width=3.25in]{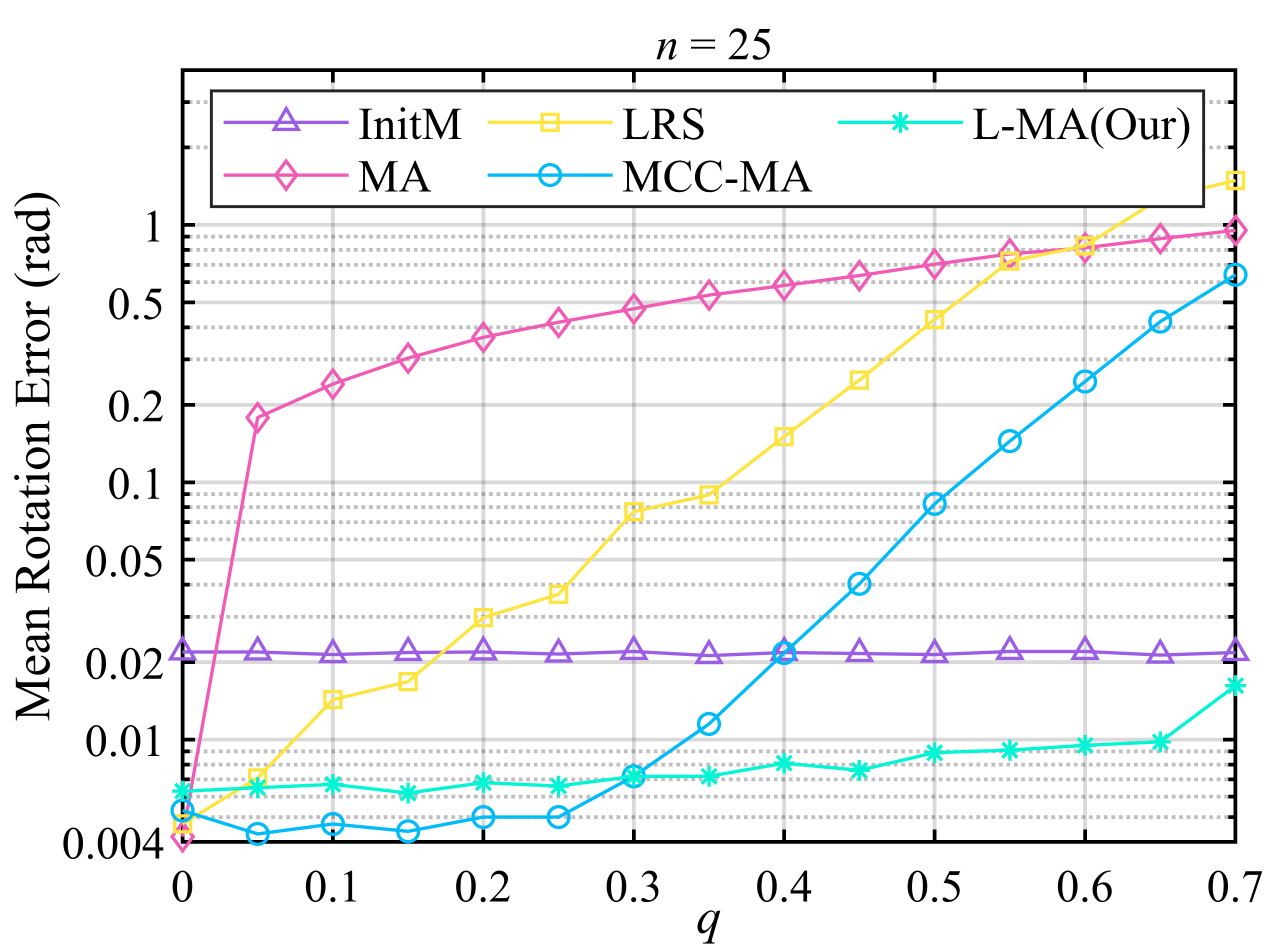} \\
      \includegraphics[width=3.25in]{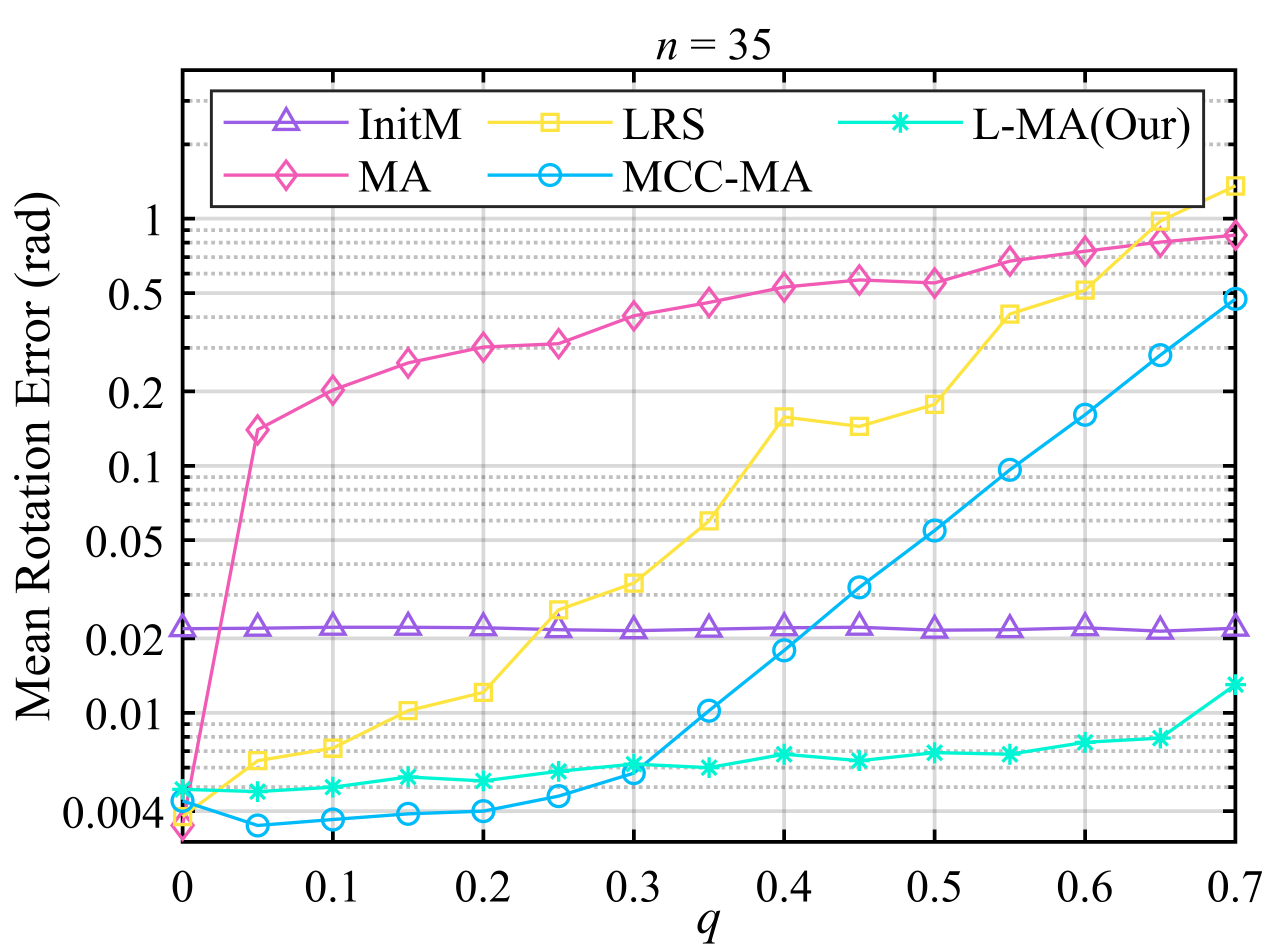} 
    \end{minipage}
  }
  \subfloat
  {  \begin{minipage}{0.55\linewidth}
      \centering
      \includegraphics[width=3.25in]{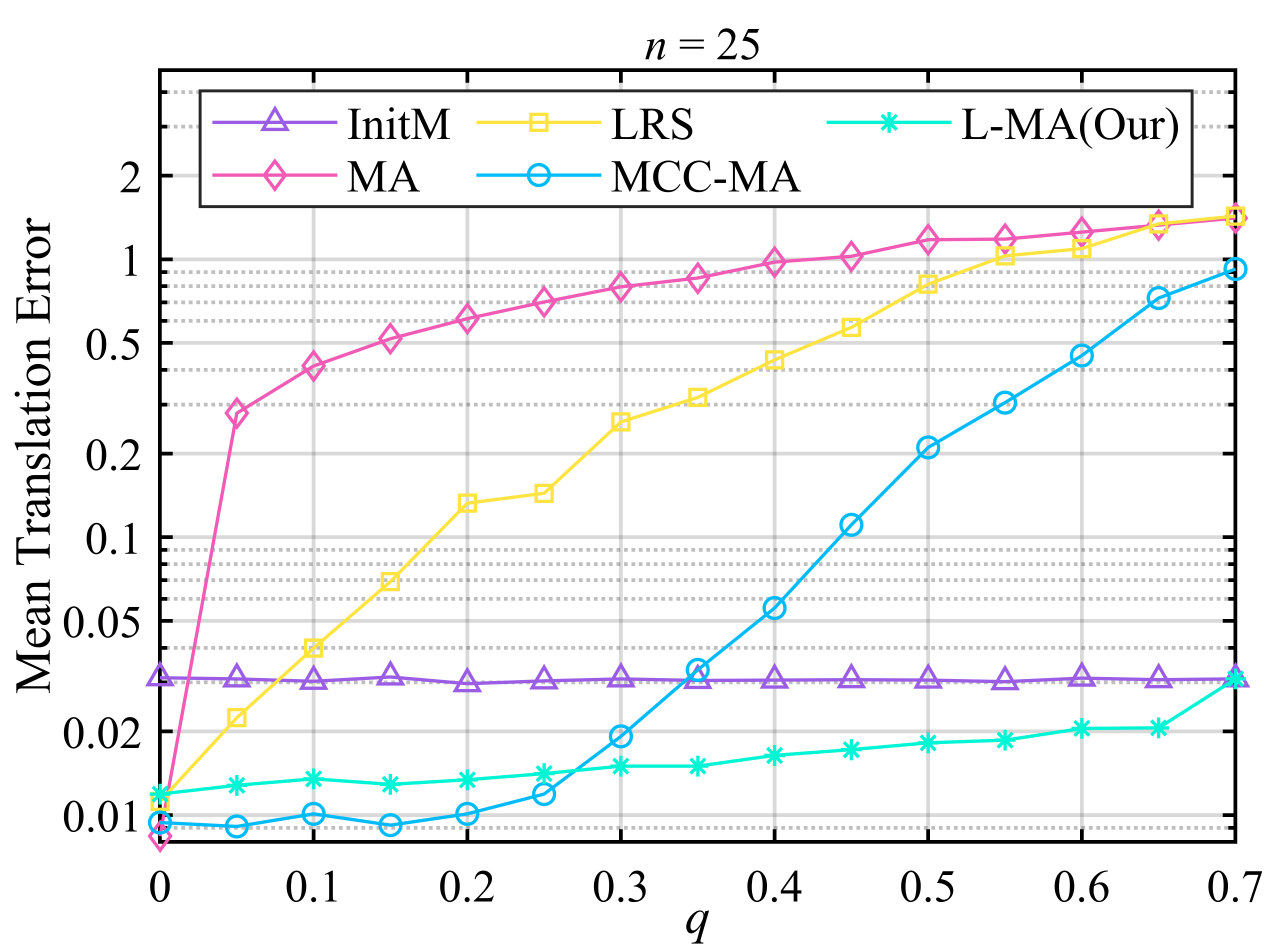} \\
      \includegraphics[width=3.25in]{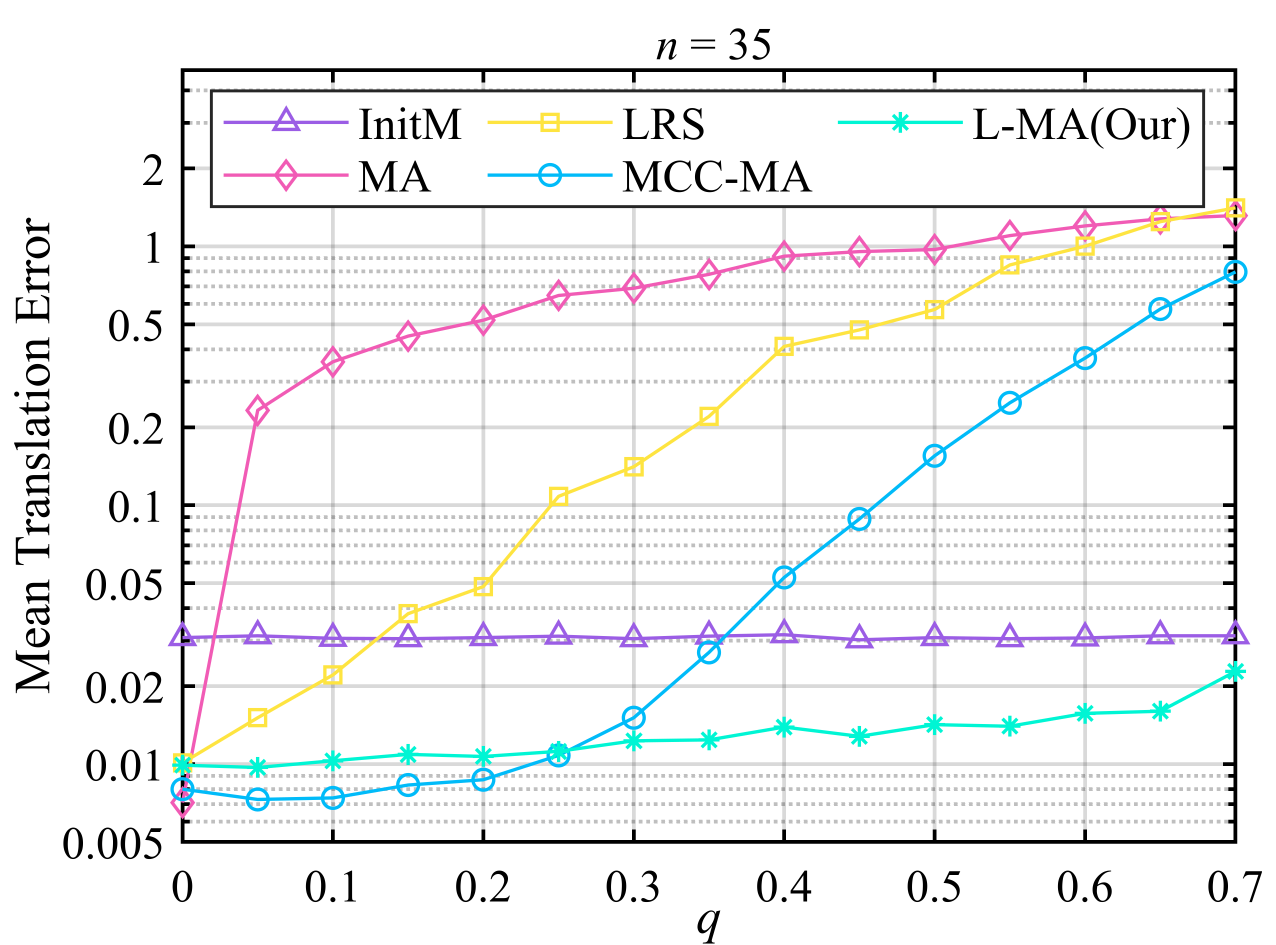} 
  \end{minipage}
  }
  \caption{Mean motion error of relative motions under different outlier ratios. The mean rotation and translation errors are displayed vs. the outlier ratio $q$, with $n=25$ (left) and $n=35$ (right). 
  Obviously, as we increase outlier ratio, the motion error of relative motions increases almost linearly.}
  \label{fig4}
  \end{figure*}

\begin{figure*}[!b]
  \centering
  \subfloat
  {
    \begin{minipage}{0.45\linewidth}
      \centering
      \includegraphics[width=3.25in]{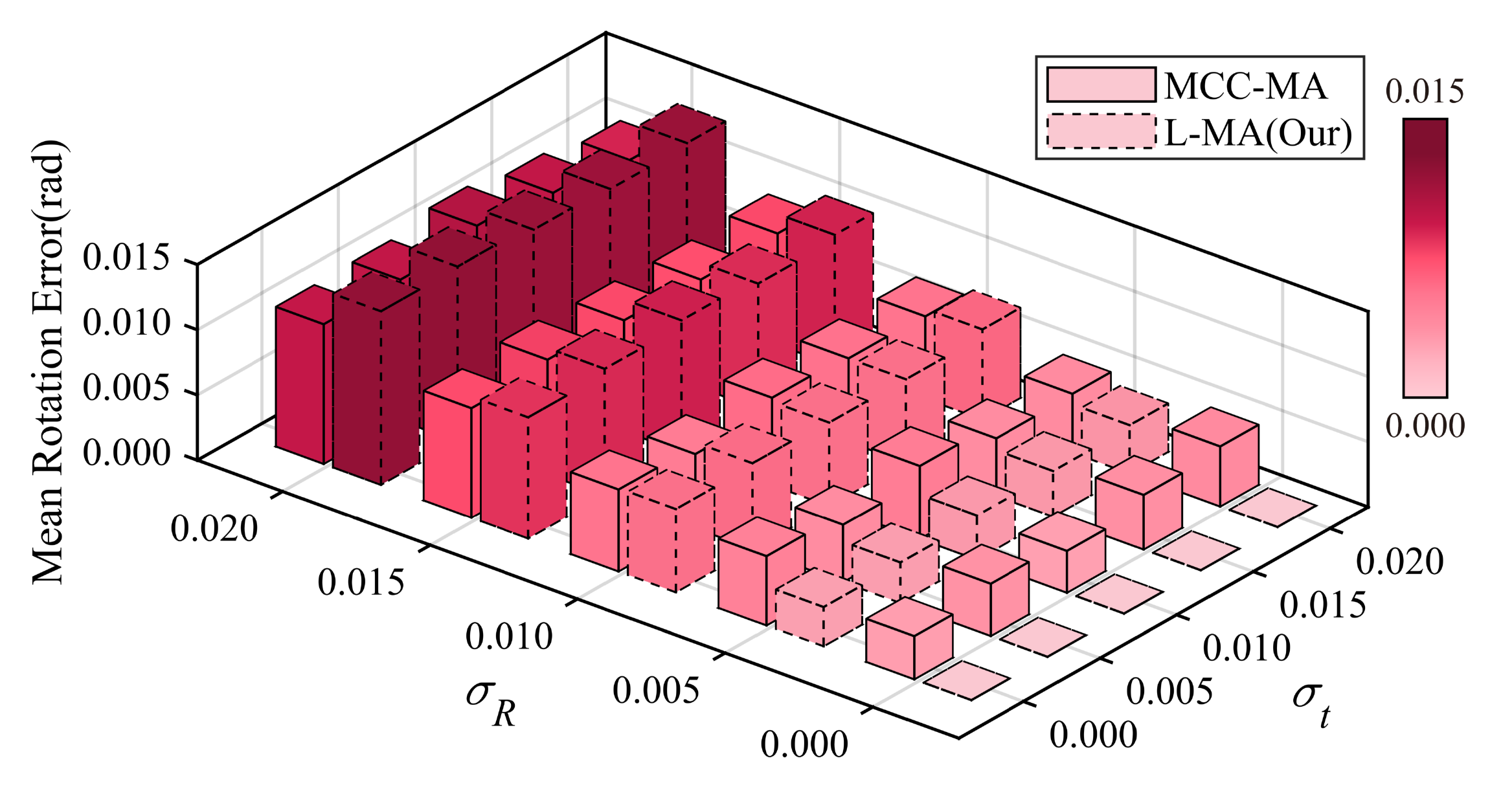} 
    \end{minipage}
  }
  \subfloat
  {  \begin{minipage}{0.55\linewidth}
      \centering
      \includegraphics[width=3.25in]{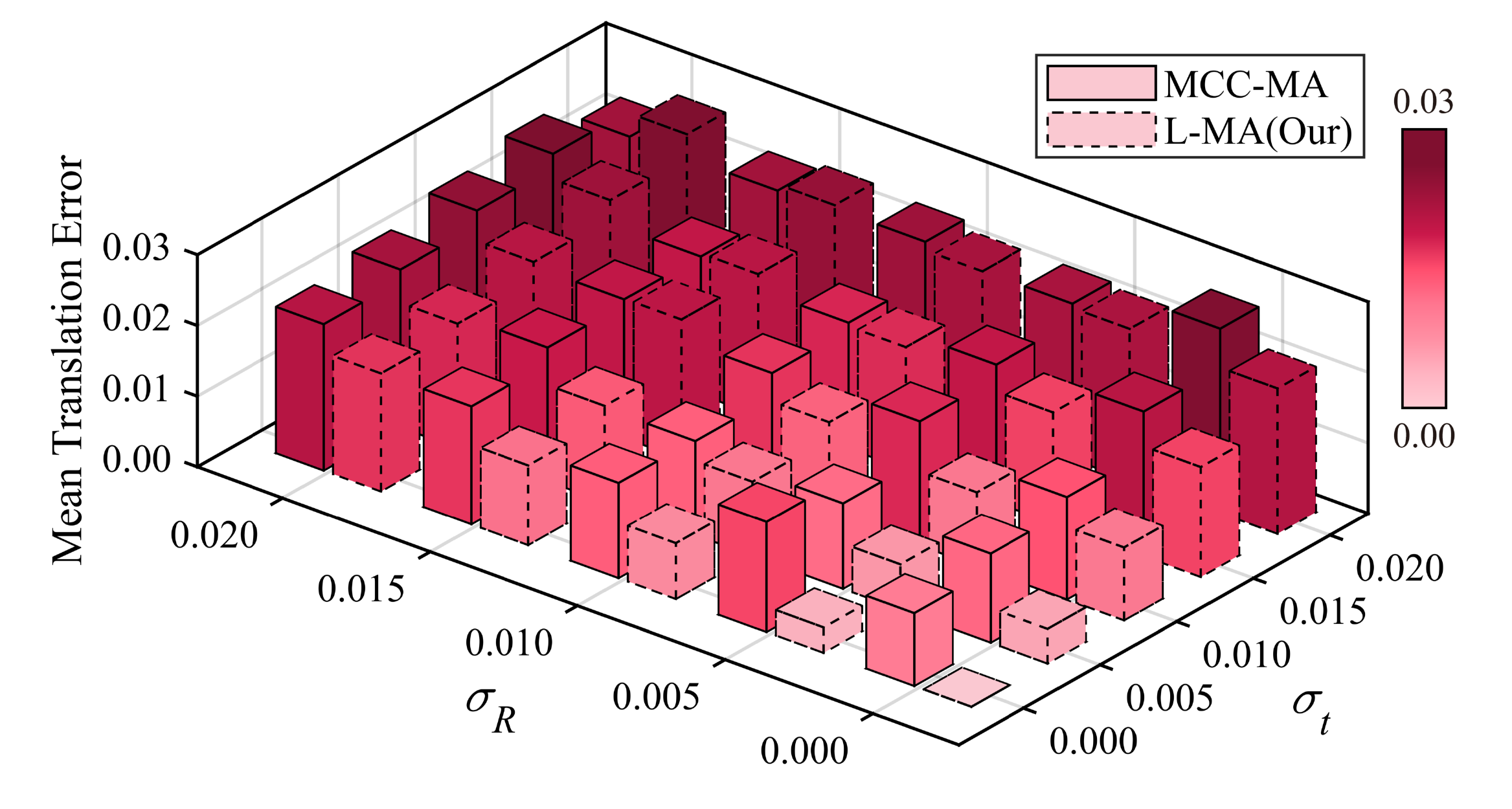} 
  \end{minipage}
  }
  \caption{Performance comparison among MCC-MA and L-MA (Our) under different noise level of inliers.
   The mean rotation and translation errors are depicted vs. noise levels of rotation and translation $\sigma _{\mathbf{R}}^{inlier}$, $\sigma _{\mathbf{t}}^{inlier}$. 
   In that figure, the results achieved by MCC-MA and our method are marked with full lines and dashed lines, respectively. }
  \label{fig5}
  \end{figure*}

Moreover, as recorded in Fig.1, LRS is the most efficient method and our method has the second fastest speed. Besides, 
MCC-MA consumes more time than MA, this is because MCC-MA needs to update the weight matrix in each iteration. In addition, 
what we observe is that an increase in the number of relative motions leads to a significant increase in runtimes of MMA and 
MA and a slight increase in that of our method and LRS. For MCC-MA and MA, a fundamental step is to calculate the pseudo-inverse 
of a large matrix whose the size of columns is the linear function of the number of relative motions, which is very time-consuming. 
Apparently, for our method, the main computation in each iteration is to solve the SOCP problem, as given in (39). However, benefiting 
from the rapid convergence speed of SOCP, the number of iterations required by our method is far less than that of 
MCC-MA so our method is more efficient than MCC-MA and its runtime usually remains close to LRS. 
In short, Our method achieves the overall best performance, in terms of accuracy and efficiency.

\subsubsection{Robustness to Outliers}

  \begin{figure*}[!b]
    \centering
    \subfloat
    {
      \begin{minipage}{0.45\linewidth}
        \centering
        \includegraphics[width=3.25in]{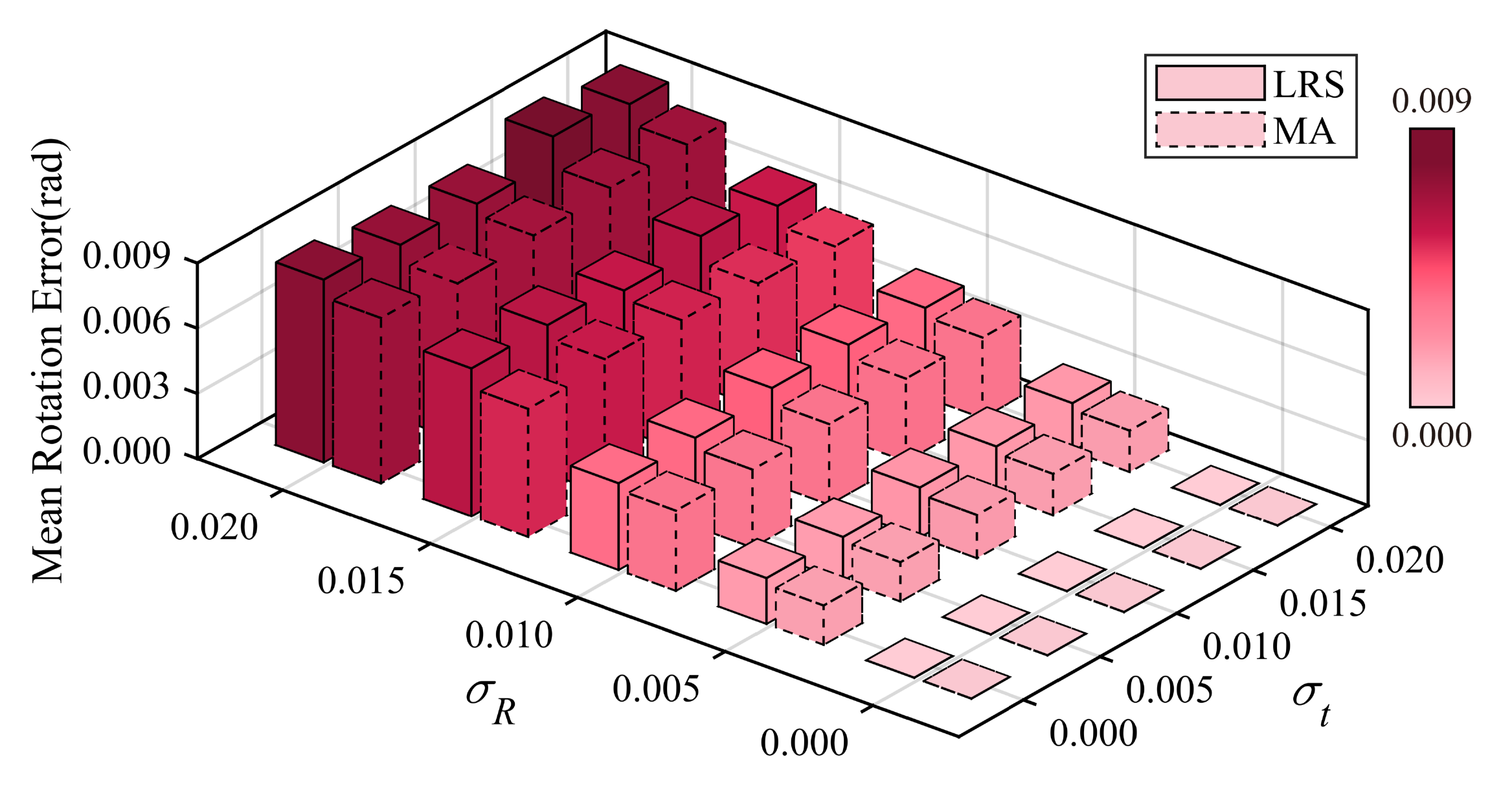} 
      \end{minipage}
    }
    \subfloat
    {  \begin{minipage}{0.55\linewidth}
        \centering
        \includegraphics[width=3.25in]{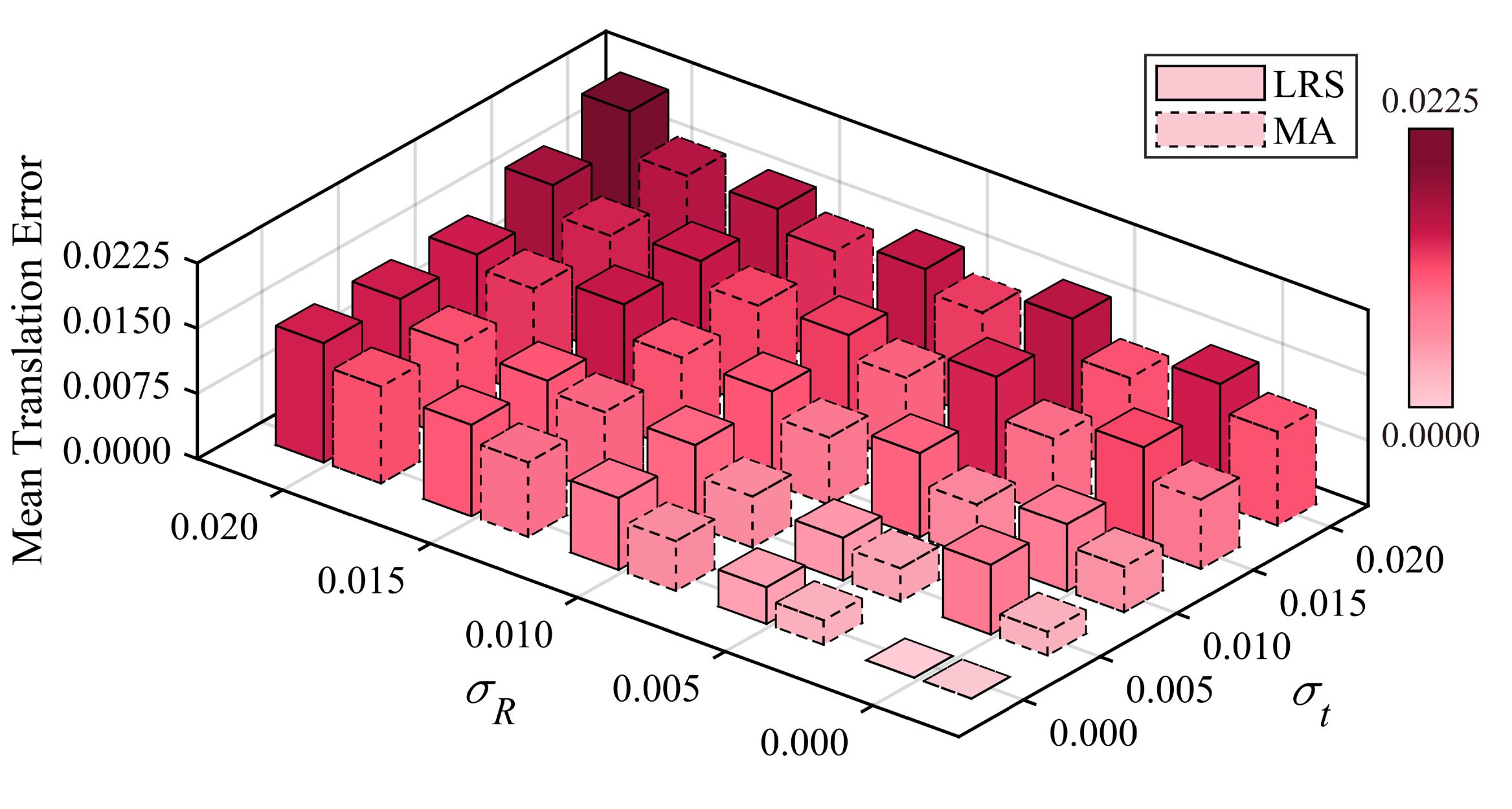} 
    \end{minipage}
    }
    \caption{Performance comparison among LRS and MA under different noise levels of inliers.
     The mean rotation and translation errors are depicted vs. noise levels of rotation and translation $\sigma _{\mathbf{R}}^{inlier}$, $\sigma _{\mathbf{t}}^{inlier}$. 
     In that figure, the results achieved by LRS and MA are marked with full lines and dashed lines, respectively. }
    \label{fig6}
    \end{figure*}

    \begin{figure*}[!b]
      \renewcommand{\dblfloatpagefraction}{.9}
        \centering
      \subfloat
      {
        \begin{minipage}{0.45\linewidth}
          \centering
          \includegraphics[width=3.25in]{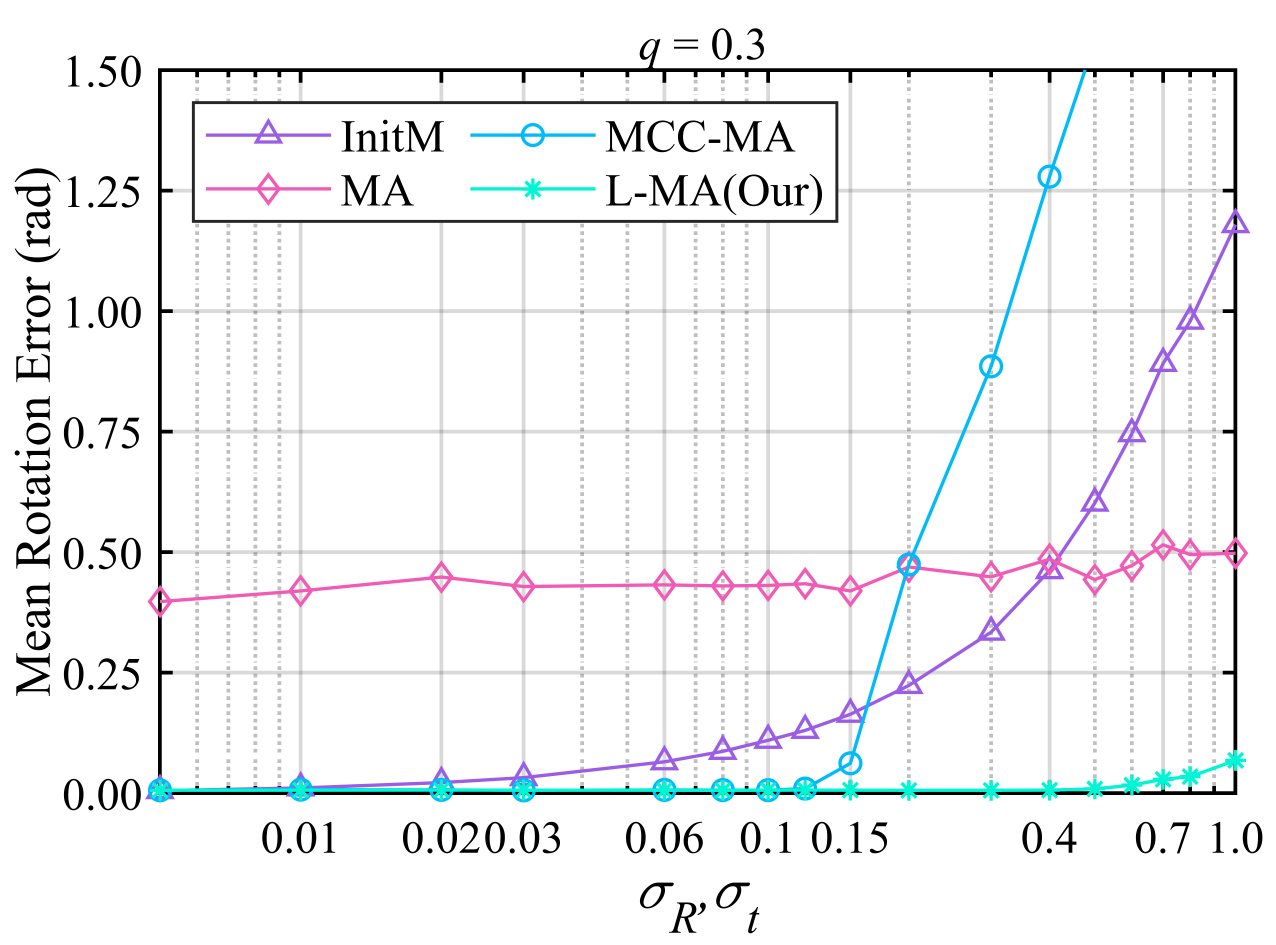} \\
          \includegraphics[width=3.25in]{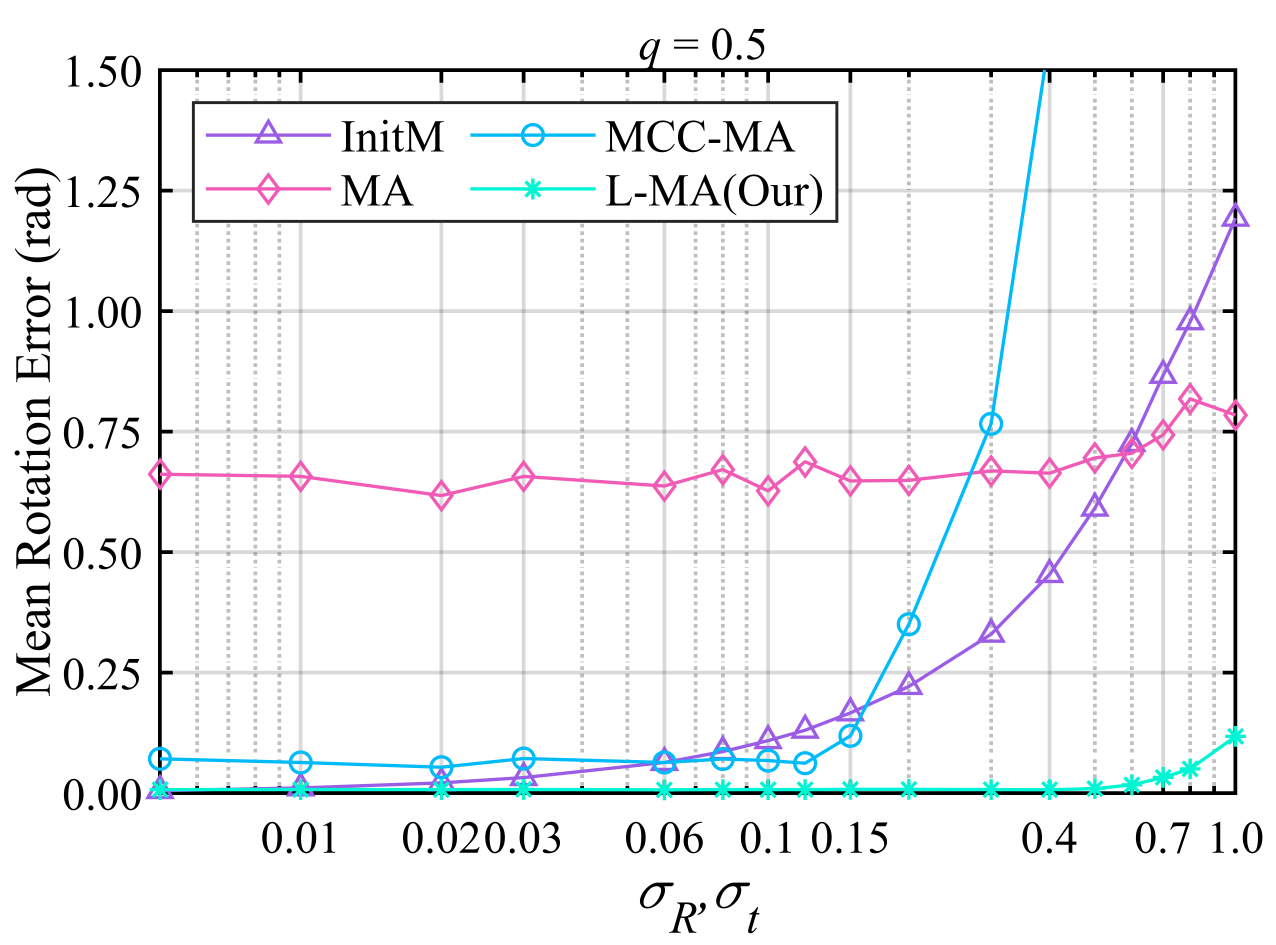} 
        \end{minipage}
      }
      \subfloat
      {  \begin{minipage}{0.55\linewidth}
          \centering
          \includegraphics[width=3.25in]{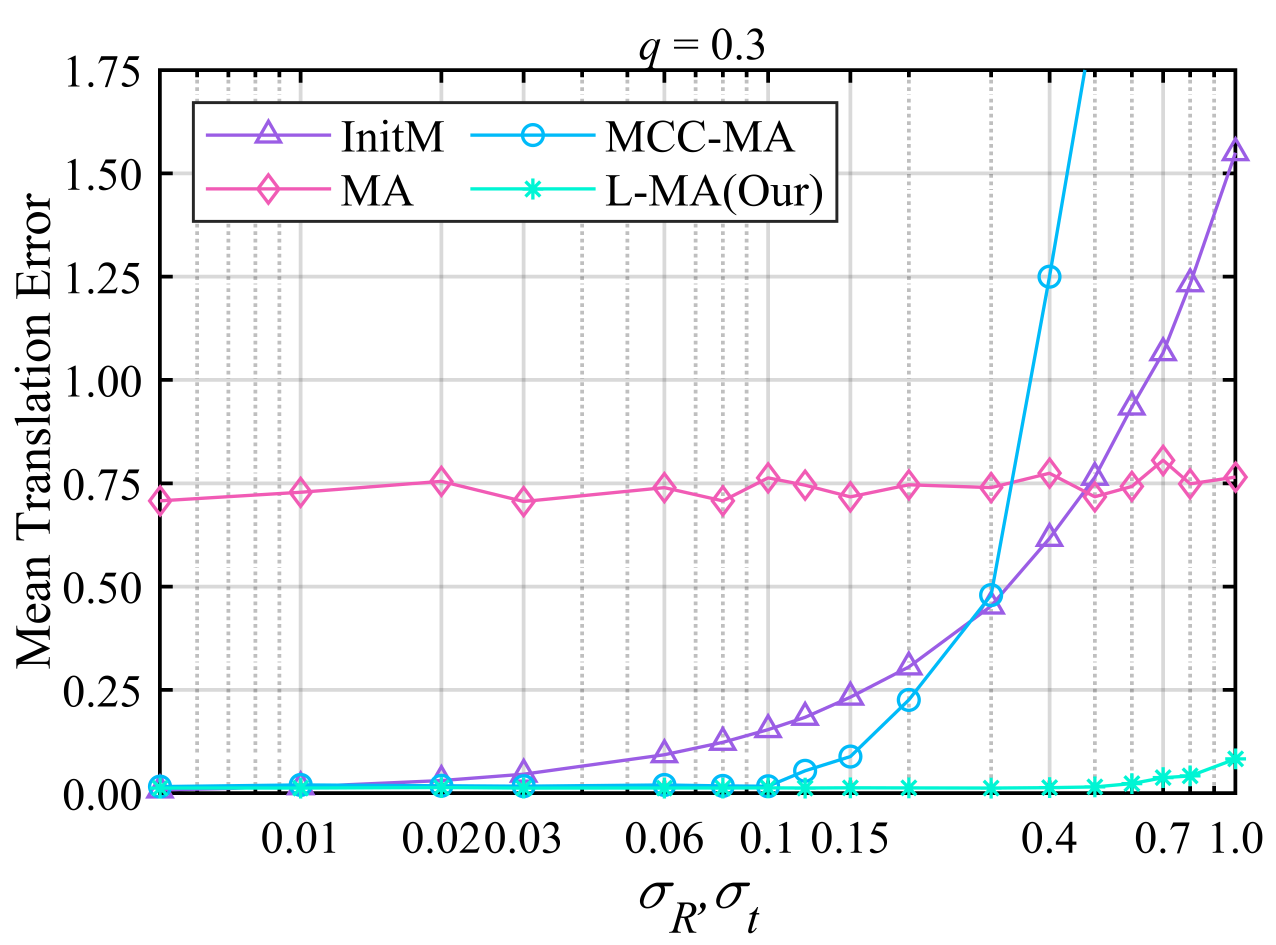} \\
          \includegraphics[width=3.25in]{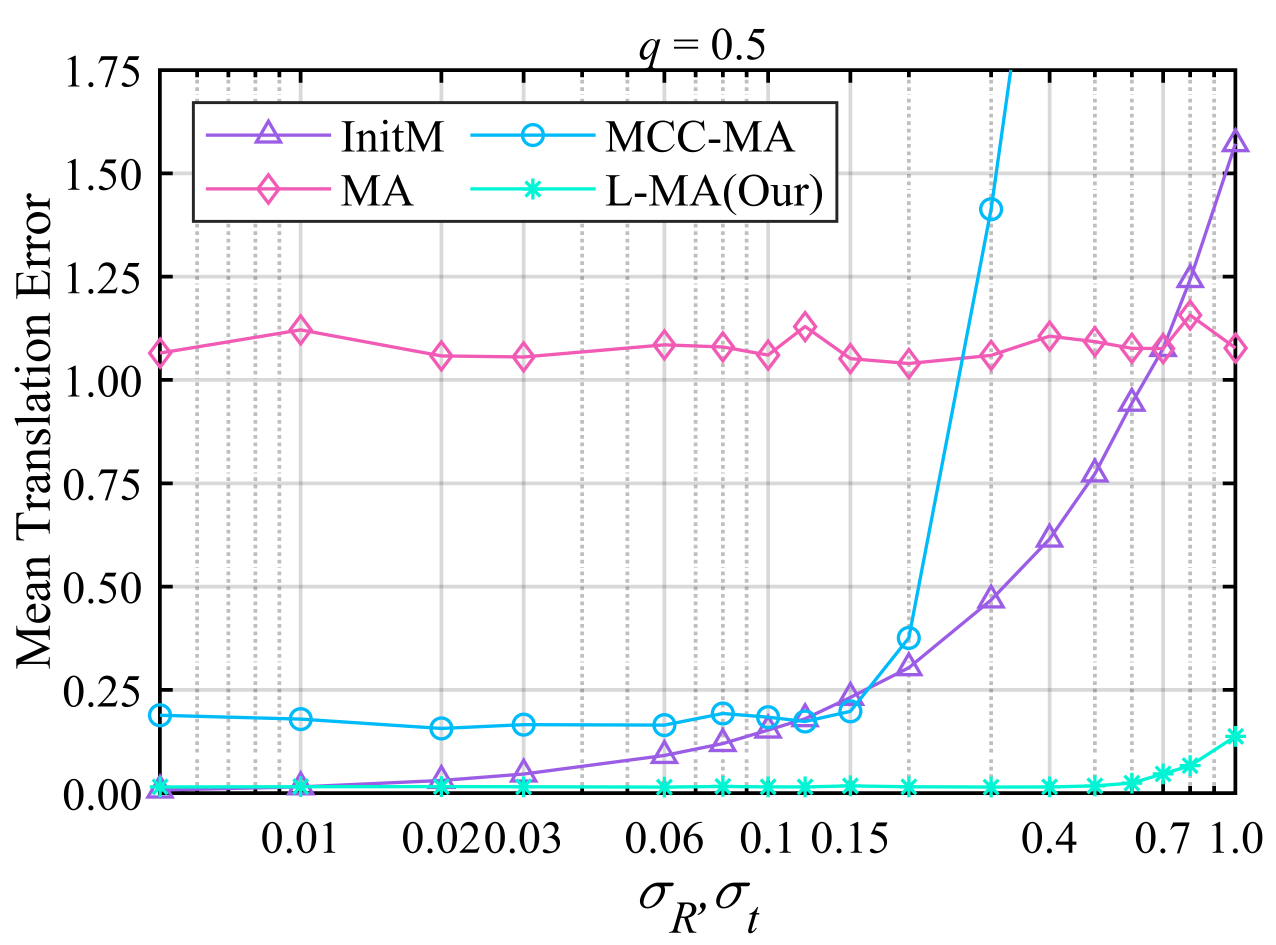} 
      \end{minipage}
      }
      \caption{
        Performance comparison among different methods under different initial motions. All mean errors are plotted vs. 
        noise levels of ground truth $\sigma _{\mathbf{R}}^{init}$, $\sigma _{\mathbf{t}}^{init}$, with $q=0.3$ (top) and $q=0.5$ (bottom). In this experiment, we apply different noises to 
        ground truth so as to obtain different initial motions. For the sake of simplicity, the noise level of rotation 
        is set equal to that of translation.
      }
      \label{fig7}
      \end{figure*}

To study the robustness to outliers of different methods, a comparative experiment is carried out with different 
values of $q$. In this experiment, the value of $q$ varies between $0.0\sim0.7$, indicating that the set of relative motions 
includes about $0\%\sim70\%$ outliers. We considered two cases: $n=25$ and $n=35$. In both cases, $\sigma _{\mathbf{R}}^{inlier}=\sigma _{\mathbf{t}}^{inlier}=0.01$,
$\sigma _{\mathbf{R}}^{init}=\sigma _{\mathbf{t}}^{init}=0.02$ and $p=0.3$.

We report the mean motion errors of relative motions as functions of $q$ in Fig.3. Moreover, we illustrate the 
results achieved by different methods in Fig.4. As depicted in Fig.3, with a increase in the value of $q$, the 
mean error of relative motions increases continuously. We have seen in Fig.4 that an increase in the outlier ratio 
leads to a decrease in the accuracy of all compared methods, the significant decrease in that of LRS and MCC-MA 
particularly. It should be noted, that our method can attain a relatively high accuracy for both rotation and 
translation even when $65\%$ of relative motions are outliers, demonstrating the strong robustness of our 
method to outliers. In addition, MA obtains, not surprisingly, the most accurate results when the outlier 
ratio is zero.

Moreover, as mentioned previously, the performances of our method and MCC-MA are affected by the kernel width. 
Instead of the mean of all Frobenius norms of residual error adopted in MCC-MA, we choose the median of residuals 
as kernel width as defined in (39), and hence, befitting from the characteristics of the median value, our 
method is more robust to outliers than MCC-MA. In addition, MCC-MA outperforms our method when the outlier 
ratio is small to $25\%$. This is because we set the value of the parameter $\alpha$, introduced in Section II.D, is 
$0.7$, indicating that the proposed method, to a certain extent, neglects the information supplied by the 
remaining $30\%$ relative motions which may lead to a slight loss of registration errors. Even so the accuracy 
of our method is still very close to that of MCC-MA under a low outlier ratio.

\subsubsection{Effect of Noises}

The performance of all compared methods is determined not only by the outlier ratio of relative motions, but 
also by the noise levels of inliers. Thus, in order to evaluate the effect of the latter, we carried out  
comparative experiments with different values of $\sigma _{\mathbf{R}}^{inlier}$, $\sigma _{\mathbf{t}}^{inlier}$. 
As both MA and LRS are poor in handling outliers, we set, respectively, the values of $q$ are $0$ for them and $0.3$ for 
MCC-MA and our method. In both experiments, we fixed the values of other parameters, i.e., $\sigma _{\mathbf{R}}^{inlier}=\sigma _{\mathbf{t}}^{inlier}=0.01$,
$p=0.3$ and $n=30$. 

Fig.5 illustrates the averaged results of MCC-MA and our method under different levels of noise and Fig.6 shows 
that of MA and LRS. As we saw in Fig.5 and Fig.6, with the increase in the noise level of inliers, the accuracies 
of all methods tested in this experiment rapidly decrease, indicating that the noise level of inliers has a 
very important influence on performances of all tested methods. Thus, we can draw the following conclusion: the 
reliability of the inliers is a critical prerequisite to obtaining excellent results for motion average algorithms. 
In addition, as displayed in Fig.5, compared to MCC-MA, our method is more sensitive to the noise level of inliers 
because (9) is satisfied as $\boldsymbol{\mathcal{J}}_{l}^{-1}\left( \boldsymbol{\xi}_{ij}^{k-1} \right)\approx \mathbf{I}$ and
$\boldsymbol{\mathcal{J}}_{r}^{-1}\left( \boldsymbol{\xi}_{ij}^{k-1} \right)\approx \mathbf{I}$  if and only if the residual  $\Delta \mathbf{M}_{ij}^{k-1}$ 
is “small”, i.e., if the noise levels of inliers are sufficiently close to zero. Fortunately, the pair-wise registration
algorithms which have high registration accuracies, such as \textit{Robust ICP} \cite{ref25} , or \textit{Sparse ICP} \cite{ref26}, may be used to avoid this drawback. In addition, a phenomenon that can be 
observed in both Fig.5 and Fig.6 is that, when the noise level of rotation remains constant, the increase in the noise level of translation does lead to the loss 
of rotational accuracy; in turn, the errors in both rotation and translation increase with noise level of rotation 
increasing when the noise level of translation remains constant.

\subsubsection{Sensibility to the Initial Global Motions}
As mentioned previously, the performances of MA, MCC-MA, and our method are dependent on the good initial motions. Thus, 
to study the sensibility to the initial global motion, experiments are carried out with different values of 
$\sigma _{\mathbf{R}}^{init}$, $\sigma _{\mathbf{t}}^{init}$. In both cases, $\sigma _{\mathbf{R}}^{inlier} = \sigma _{\mathbf{t}}^{init}=0.01$
and $n=30$. Specially, we tested the algorithms under different outlier ratios, i.e., $q=0.3$ and $q=0.5$. Fig.7 shows the mean errors of 
results obtained by different methods as a function of $\sigma _{\mathbf{R}}^{init}$, and of $\sigma _{\mathbf{t}}^{init}$.

As illustrated in Fig.7, MCC-MA is difficult to recover global motion when the noise level of initialization exceeds 0.15 and 
the performance of MCC-MA rapidly decreases as we continuously increase the noise level of initialization. On the 
on hand, the pseudo-inverse of $\boldsymbol{J}$ that MCC-MA needed to determine in each iteration may cause a convergence problem 
and further lead to a failure of multi-view registration because it is easy to be singular under poor initializations. 
On the other hand, poor initializations may lead to the kernel width for MCC-MA in the first iteration so large as to 
assign unreliable weight for each relative motion. Compared to MCC-MA, our method always converges to good global 
motions, even though the noise level of initialization is $0.4$. What this implies is that our method has the low sensibility 
to the initialization and a large convergence domain.

\subsection{Real Data}

\begin{table}[!b]
  \caption{Statistics Information of Relative Motions under \\
    $30\%$ Overlap Percentage Threshold}
  \renewcommand{\arraystretch}{1.3}
  \centering
    \begin{tabular}{cccccc}
    \toprule
    \multicolumn{1}{c}{\multirow{2}{*}{Dataset}} & \multicolumn{1}{c}{\multirow{2}{*}{Motions}}  & \multicolumn{2}{c}{Rotation error(rad)} & \multicolumn{2}{c}{Translation error(mm)} \\
    \cline{3-6} 
    & &Mean&Meadian&Mean&Meadian \\
    \midrule
    Armadillo	&57&	0.0478&	0.0042&	5.5610&	0.3409 \\
    Buddha&	96&	0.2200	& 0.0043	&2.1814 &	0.4097 \\ 
    Bunny&	45&	0.0789	&0.0025&	3.0070&	0.2889 \\ 
    Dragon&	77&	0.2786&	0.0041&	20.839&	0.3641\\ 
    \bottomrule
    \end{tabular}
    \label{tab1}
  \end{table}
  
  \begin{table}[!b]
  \caption{Statistics Information of Relative Motions under \\
    $20\%$ Overlap Percentage Threshold}
  \renewcommand{\arraystretch}{1.3}
  \centering
    \begin{tabular}{cccccc}
    \toprule
    \multicolumn{1}{c}{\multirow{2}{*}{Dataset}} & \multicolumn{1}{c}{\multirow{2}{*}{Motions}}  & \multicolumn{2}{c}{Rotation error(rad)} & \multicolumn{2}{c}{Translation error(mm)} \\
    \cline{3-6} 
    & &Mean&Meadian&Mean&Meadian \\
    \midrule
    Armadillo&	81&	0.3724	&0.0057&	15.015&	0.4907 \\
    Buddha&	131&	0.6428	&0.0059	&5.6311	&0.5352 \\ 
    Bunny&	64&	0.5363	&0.0045&	28.069	&0.4160 \\
    Dragon&	115&	0.7042&	0.0080&	53.874&	0.5778\\
    \bottomrule
    \end{tabular}
    \label{tab2}
  \end{table}
  
\begin{table*}[!t]
  \caption{Performance Comparison among Different Multi-view Registration Methods \\
  on Four Data Sets under $30\%$ Overlap Percentage Threshold
  }
  \renewcommand{\arraystretch}{1.3}
  \centering
  \resizebox{0.98\textwidth}{!}{
    \begin{tabular}{c||cc|cc||cc|cc||cc|cc||cc|cc||}
    \toprule
    \multicolumn{1}{c||}{\multirow{2}{*}{Methods}} & \multicolumn{4}{c||}{Armadillo} & \multicolumn{4}{c||}{Buddha} & \multicolumn{4}{c||}{Bunny}& \multicolumn{4}{c||}{Dragon} \\
    \cline{2-17}
    &${{e}_{\mathbf{R}}}$(rad)&${{e}_{\mathbf{t}}}$mm&T(s)&Iter&${{e}_{\mathbf{R}}}$(rad)&${{e}_{\mathbf{t}}}$mm&T(s)&Iter&${{e}_{\mathbf{R}}}$(rad)&${{e}_{\mathbf{t}}}$mm&T(s)&Iter&${{e}_{\mathbf{R}}}$(rad)&${{e}_{\mathbf{t}}}$mm&T(s)&Iter\\
    \midrule
    InitM	&0.0269	&2.5680&	$/$	&$/$	&0.0207	&1.3401	&$/$&$	/$&	0.0255&	3.4404&	$/$	&$/$&	0.0241&	5.8683&	$/$&	$/$\\
    LRS&	0.0077&	1.1133&	$\mathbf{0.0091}$&	$/$	&0.0117&	0.6307&	$\mathbf{0.0171}$&$	/$	&0.0375&	7.8921&	$\mathbf{0.0133}$&	$/$	&0.6280	&10.6777	&$\mathbf{0.0169}$&	$/$ \\
    MA&	0.0854&	12.5369	&0.3685&	66&	0.1592&	1.3574&	1.0950&	110	&0.2232&	13.4495	&0.3522&	78	&0.4893	&51.2763&	1.1821&	130 \\
    WMAA&	0.0097	&1.5094	&0.3273	&66	&0.0191&	0.6612&	0.7586	&83&	0.0334	&2.3219	&0.2489&	59&	0.2475&	28.5714&	0.9293	&120\\
    MCC-MA	&$\mathbf{0.0053}$	&0.7094&	0.3566&	52&	0.0084&	0.6484&	0.7879	&62	&0.0041&	0.3480&	0.2807&	51	&0.0121	&0.9734&0.8759&	87  \\
    L-MA(Our)&	0.0072&	$\mathbf{0.6838}$&	0.2077&	$\mathbf{8}$&	$\mathbf{0.0062}$&	$\mathbf{0.5659}$	&0.2605&	$\mathbf{7}$	& $\mathbf{0.0027}$&	$\mathbf{0.2308}$&	0.1486	&$\mathbf{6}$	& $\mathbf{0.0089}$	&$\mathbf{0.7571}$&	0.3889& $\mathbf{12}$\\
    \bottomrule
    \end{tabular}}
    \label{tab3}
  \end{table*}
  
  \begin{figure*}[htbp]
    \renewcommand{\dblfloatpagefraction}{.9}
      \centering
    \subfloat
    {
      \begin{minipage}{0.10\linewidth}
        \flushleft
        \includegraphics[height=3.5in]{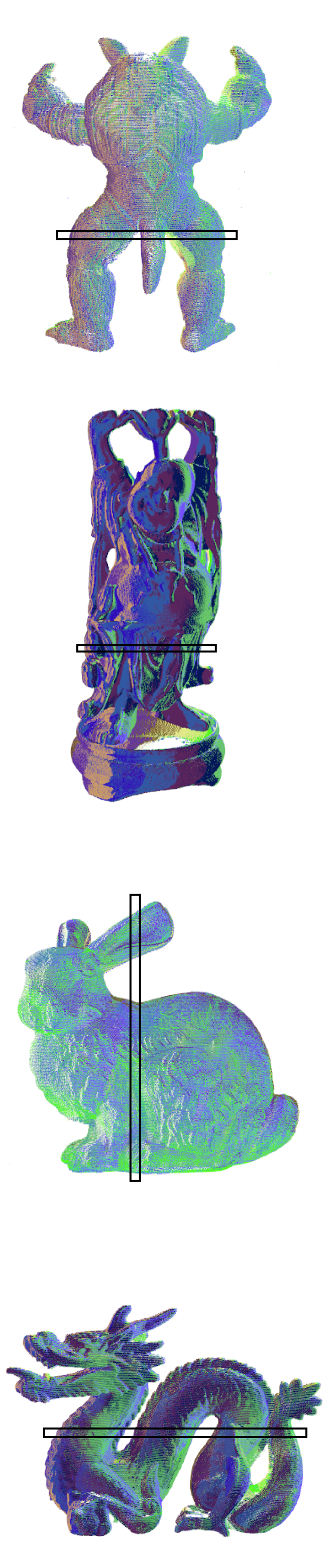} 
        \centerline{(a)}
      \end{minipage}
    }
    \subfloat
    {  \begin{minipage}{0.13\linewidth}
        \flushleft
        \includegraphics[height=3.5in]{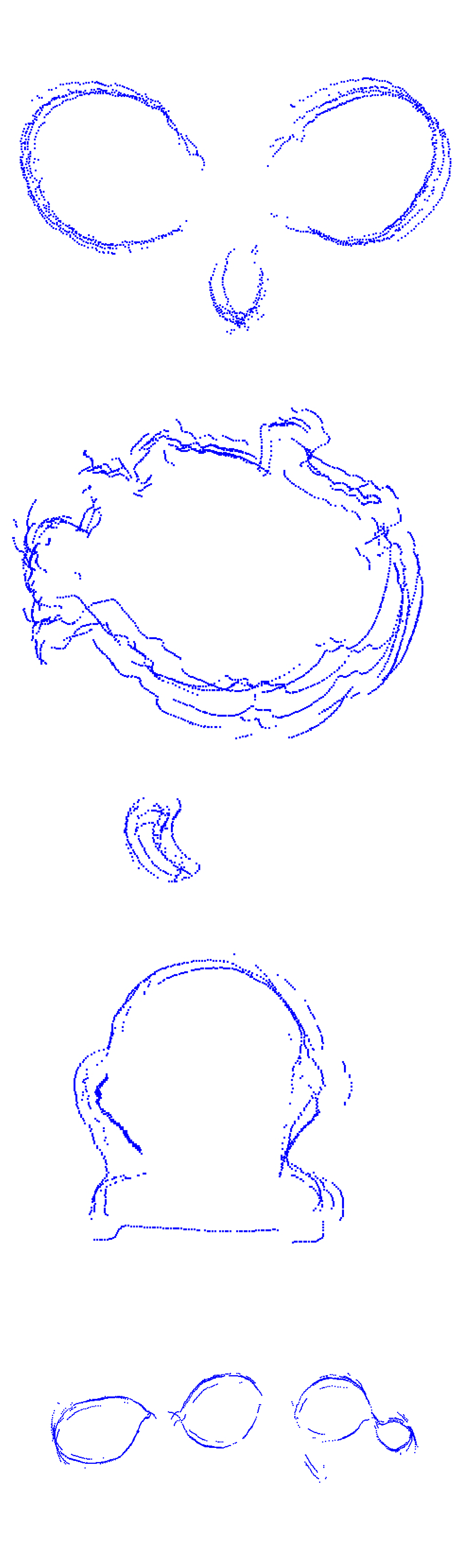} 
        \centerline{(b)}
    \end{minipage}
    }
    \subfloat
    {  \begin{minipage}{0.14\linewidth}
        \flushleft
        \includegraphics[height=3.5in]{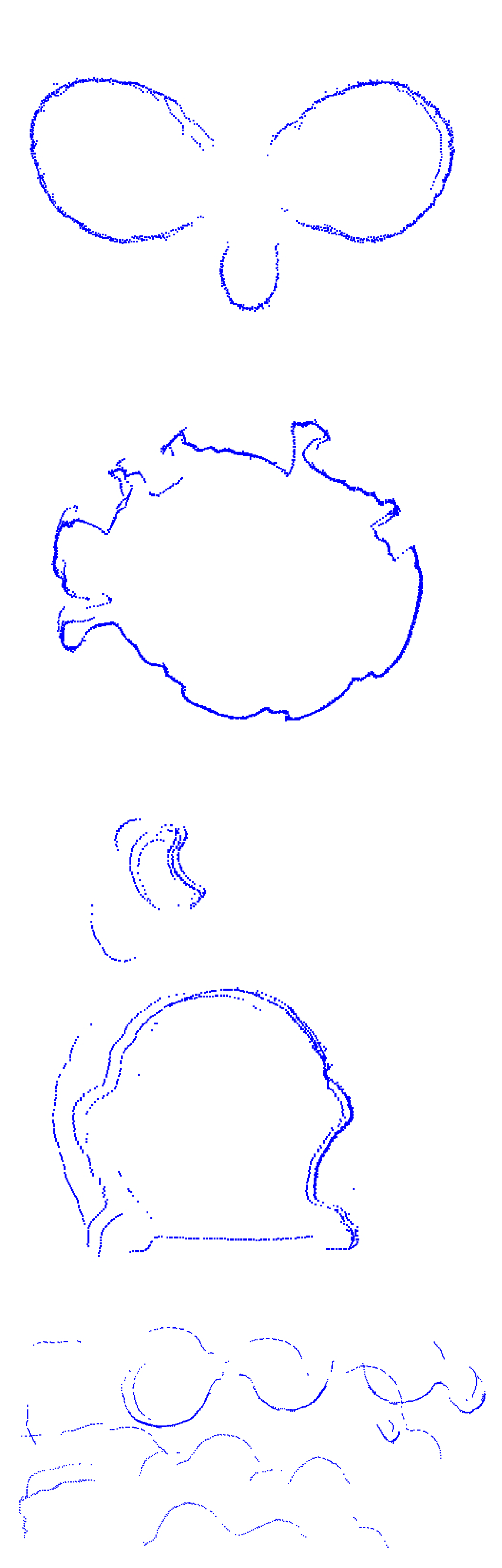} 
        \centerline{(c)}
    \end{minipage}
    }
    \subfloat
    {  \begin{minipage}{0.13\linewidth}
        \centering
        \includegraphics[height=3.5in]{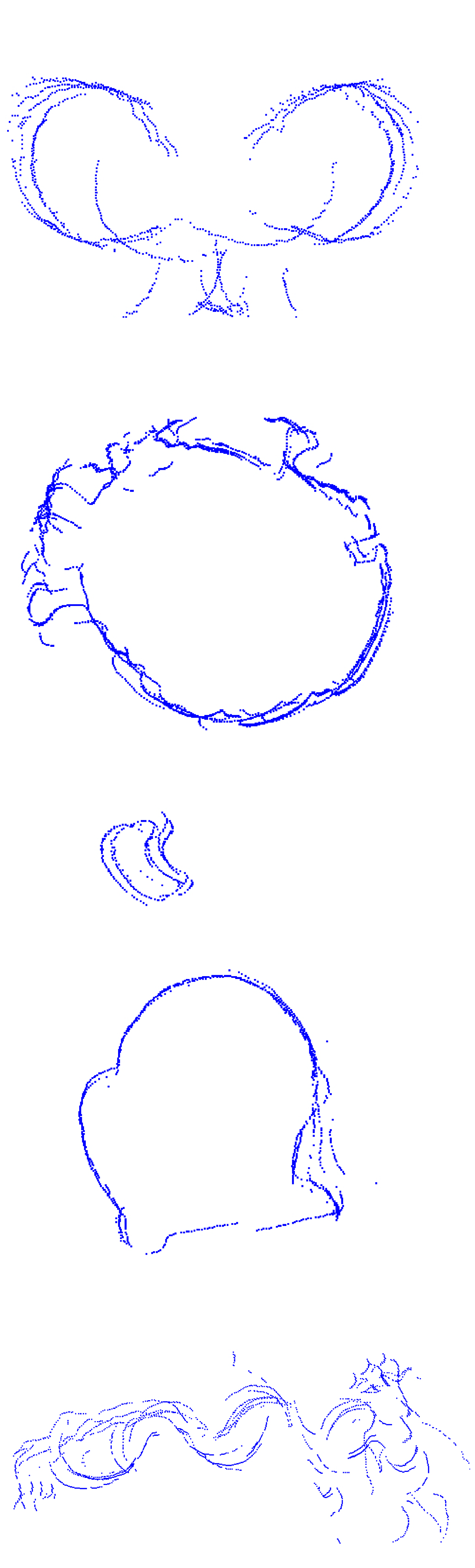} 
        \centerline{(d)}
    \end{minipage}
    }
    \subfloat
    {  \begin{minipage}{0.12\linewidth}
        \flushleft
        \includegraphics[height=3.5in]{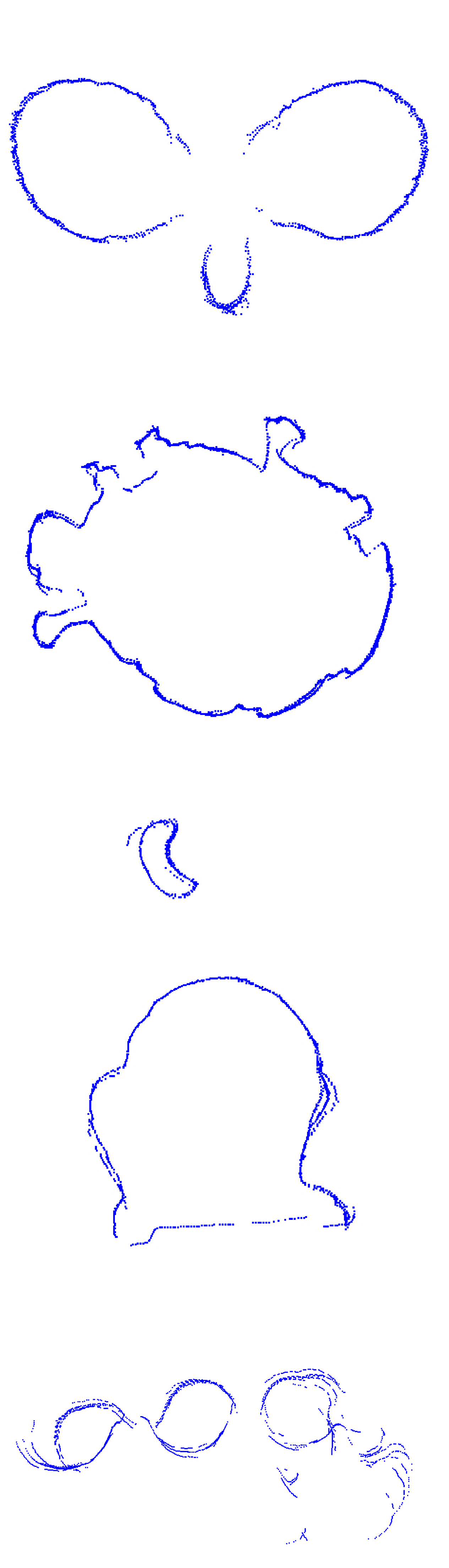}
        \centerline{(e)} 
    \end{minipage}
    }
    \subfloat
    {  \begin{minipage}{0.13\linewidth}
        \flushleft
        \includegraphics[height=3.5in]{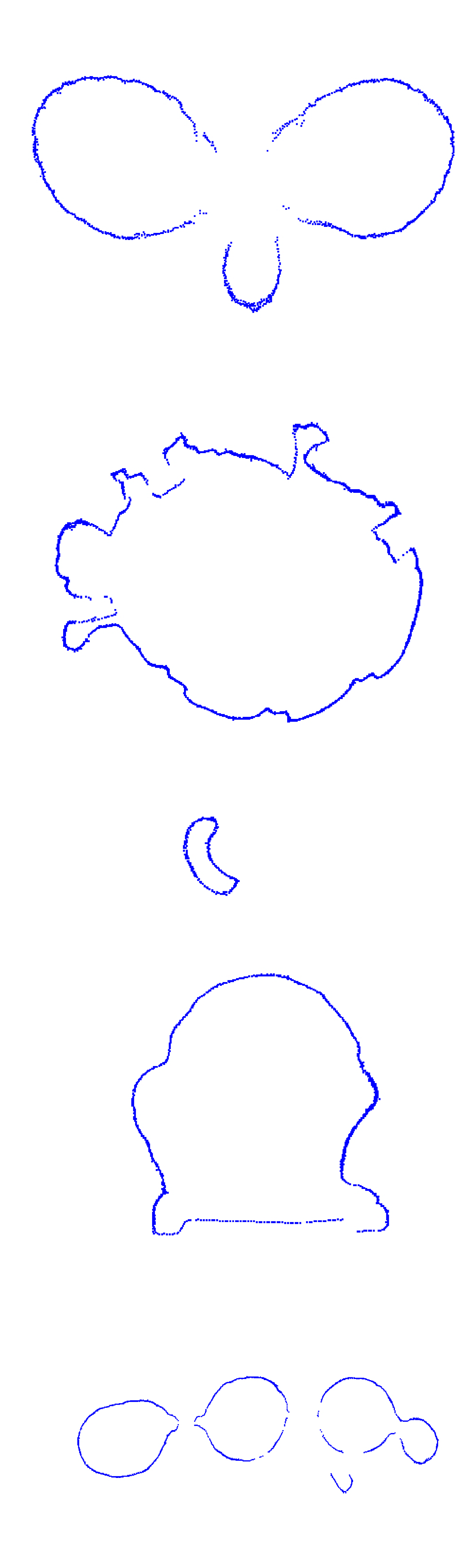} 
        \centerline{(f)}
    \end{minipage}
    }
    \subfloat
    {  \begin{minipage}{0.13\linewidth}
        \flushleft
        \includegraphics[height=3.5in]{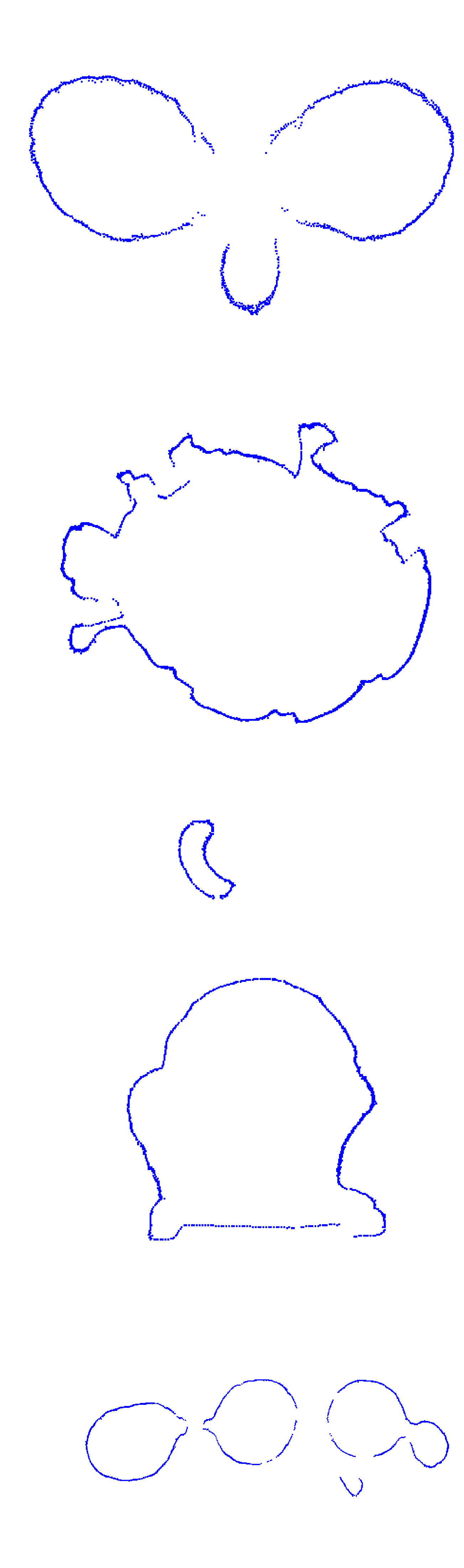} 
        \centerline{   (g)}
    \end{minipage}
    }
    \caption{
      Multi-view registration results of different methods presented in the form of cross-section for 
      four data sets under $30\%$ overlap percentage threshold: (a) Ground truth models. (b) Initial results. (c) LRS results. (d) MA results. 
      (e) WMAA results. (f) MCC-MA results. (g) Our results.
    }
    \label{fig8}
    \end{figure*}

In this experiment, we compare the performance of different methods on four benchmark data sets, which are taken from the 
Stanford 3D Scanning Repository \cite{ref30} (Bunny, Buddha, Dragon, and Armadillo, which include $10$, $15$, $15$, and $12$ scans, respectively). 
For these data sets, the ground truths of global motions are available. Note that ground truths are only used for the evaluation of 
registration results and the input relative motions. To demonstrate the performance, our method is compared with four methods 
for motion averaging, including LRS, MA, MCC-MA, and WMAA. Since all compare methods take relative motions as input and the 
pair-wise registration algorithms are sensitive to partial overlaps, we adopted the method proposed in \cite{ref16} to roughly 
estimate the overlapping percentage for each scan pair, and then utilize the pair-wise registration algorithm introduced in \cite{ref7}
to obtain relative motions for these scans pairs whose overlapping percentage are large than a pre-set threshold $\gamma_{thr}$. 
Moreover, we utilized feature-based registration \cite{ref31} to estimate initial global motions for multi-view registration. To illustrate 
the performance of our method, we test the algorithms with different overlapping percentages, namely, $\gamma_{thr}=0.3$ and $\gamma_{thr}=0.2$.
We evaluate the mean, and median error of the corresponding relative motions under these overlapping percentages.

The errors as well as the number of relative motions for $\gamma_{thr}=0.3$ and $\gamma_{thr}=0.2$, are summarized in Table I and Table II, respectively. Table III and Table IV record the registration results achieved 
by different methods. The column 'Iter' indicates the number of iterations required by different methods for convergence and we did not report that 
of LRS because it is not an iterative algorithm. To visualize the registration results, we further provide the results of different methods presented 
in the form of cross-sections for $\gamma_{thr}=0.3$ and $\gamma_{thr}=0.2$ in Fig.8 and Fig.9, respectively.

As shown in Table I and Table II, relative motions contain many outliers and the outlier ratios increase with the decrease of 
the overlapping percentage threshold. Furthermore, as illustrated in Table III and Table IV, our method achieves the lowest mean 
rotation and translation on these data sets with different overlapping percentages, while MCC-MA only outperforms our method for 
the rotation estimation on Armadillo ($30\%$ threshold). As discussed in Subsection IV.A(1), LRS is the most efficient method but it fails 
to return accurate estimates of global motions for these relative motions whose outlier ratios are relatively high, such as Dragon, and
Buddha. In addition, as reported on Table III and Table IV, the number of iterations of our method is nearly 10 times less than 
that of MA, WMAA, and MCC-MA because of the high convergence speed of our method. For this reason, our method is more efficient than MA, WMAA, and MCC-MA.

\begin{table*}[!t]
  \caption{Performance Comparison among Different Multi-view Registration Methods \\
   on Four Data Sets under $20\%$ Overlap Percentage Threshold}
  \renewcommand{\arraystretch}{1.3}
  \centering
  \resizebox{0.98\textwidth}{!}{
    \begin{tabular}{c||cc|cc||cc|cc||cc|cc||cc|cc||}
    \toprule
    \multicolumn{1}{c||}{\multirow{2}{*}{Methods}} & \multicolumn{4}{c||}{Armadillo} & \multicolumn{4}{c||}{Buddha} & \multicolumn{4}{c||}{Bunny}& \multicolumn{4}{c||}{Dragon} \\
    \cline{2-17}
    &${{e}_{\mathbf{R}}}$(rad)&${{e}_{\mathbf{t}}}$mm&T(s)&Iter&${{e}_{\mathbf{R}}}$(rad)&${{e}_{\mathbf{t}}}$mm&T(s)&Iter&${{e}_{\mathbf{R}}}$(rad)&${{e}_{\mathbf{t}}}$mm&T(s)&Iter&${{e}_{\mathbf{R}}}$(rad)&${{e}_{\mathbf{t}}}$mm&T(s)&Iter\\
    \midrule
    InitM	&0.0269	&2.5680&	$/$	&$/$	&0.0207	&1.3401	&$/$&$	/$&	0.0255&	3.4404&	$/$	&$/$&	0.0241&	5.8683&	$/$&	$/$\\
    LRS&	0.0318&	2.9198&	$\mathbf{0.0114}$&	$/$	&1.5279&	7.9560&	$\mathbf{0.0192}$&	$/$	&0.0834&	8.2177&$\mathbf{0.0147}$	&$/$	&0.3576&	21.8699&	$\mathbf{0.0188}$&	$/$ \\
    MA&	0.2679&	24.6564&	0.7799&	79	&0.1259	&2.7785	&1.7219	&109&	0.4636&	39.7073&	0.7098&	81	&0.5851&	54.4524	&2.2276	&142 \\
    WMAA&	0.0286&	2.1120	&0.5061&	61&	0.0242	&0.7471&	1.2842&	86&	0.0488&	4.0199	&0.3246	&48&	0.2992&	33.2293	&1.6596	&126 \\
    MCC-MA&	0.0211&	0.9076&	0.7553&	67&0.0797&0.7101	&2.0911&	101	&0.0080&	0.6721&	0.4441&	50&	0.0387&	1.3876&	1.5368&	90 \\
    L-MA(Our)&$\mathbf{0.0055}$&	$\mathbf{0.6947}$&	0.1653&	$\mathbf{5}$&	$\mathbf{0.0058}$&	$\mathbf{0.5668}$	&0.3622&	$\mathbf{8}$	& $\mathbf{0.0019}$&	$\mathbf{0.2508}$&	0.1748	&$\mathbf{6}$	& $\mathbf{0.0091}$	&$\mathbf{0.7422}$&	0.4552	& $\mathbf{12}$ \\
    \bottomrule
    \end{tabular}}
    \label{tab4}
  \end{table*}
  
  \begin{figure*}[htbp]
    \renewcommand{\dblfloatpagefraction}{.9}
      \centering
    \subfloat
    {
      \begin{minipage}{0.10\linewidth}
        \flushleft
        \includegraphics[height=3.5in]{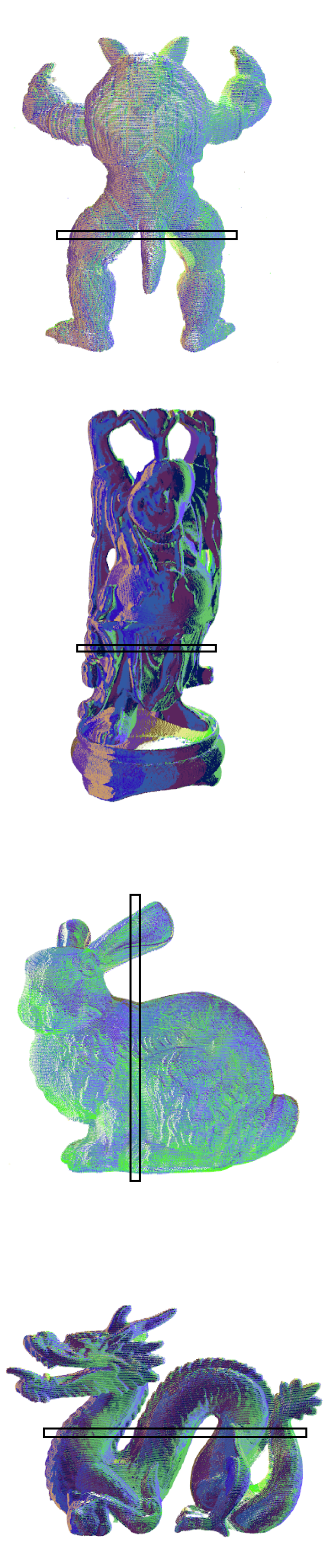} 
        \centerline{(a)}
      \end{minipage}
    }
    \subfloat
    {  \begin{minipage}{0.13\linewidth}
        \flushleft
        \includegraphics[height=3.5in]{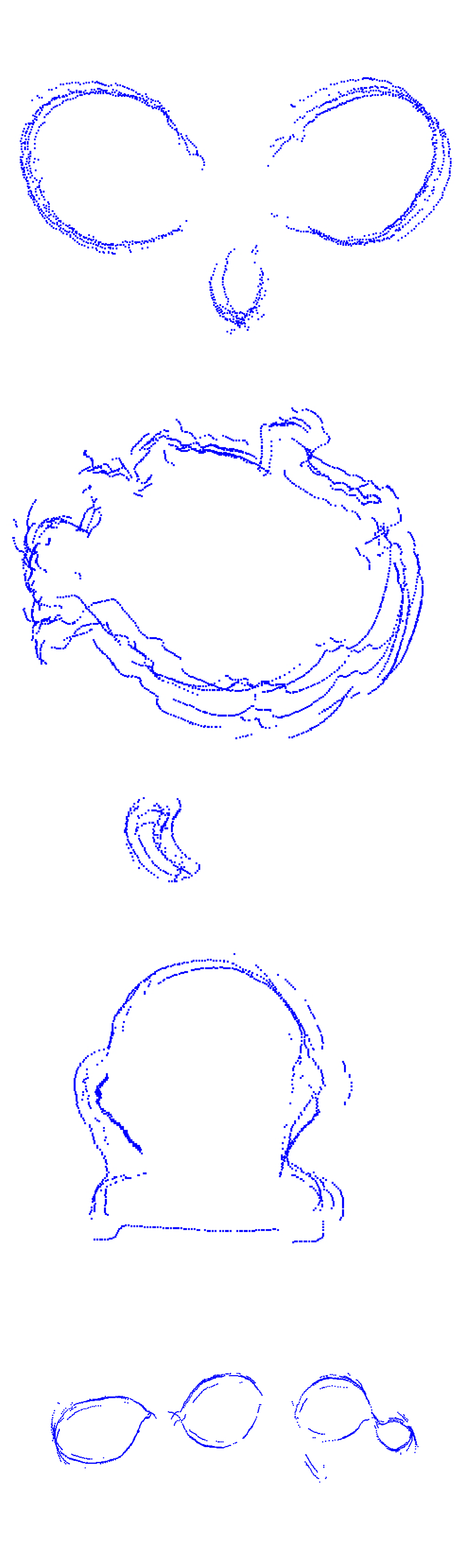} 
        \centerline{(b)}
    \end{minipage}
    }
    \subfloat
    {  \begin{minipage}{0.13\linewidth}
        \flushleft
        \includegraphics[height=3.5in]{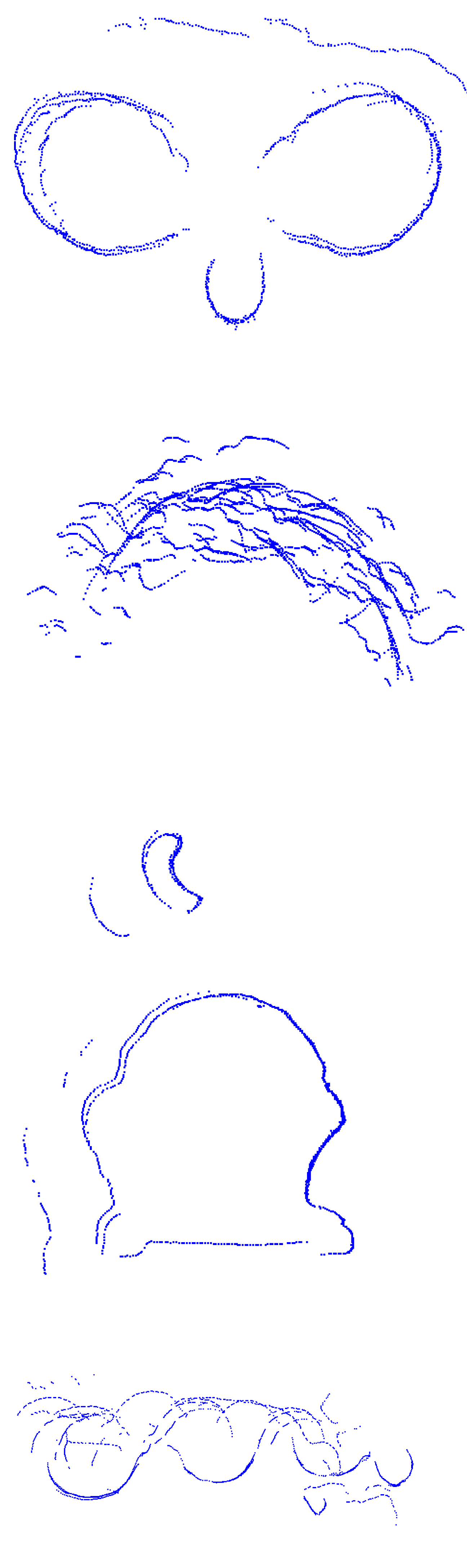} 
        \centerline{(c)}
    \end{minipage}
    }
    \subfloat
    {  \begin{minipage}{0.125\linewidth}
        \centering
        \includegraphics[height=3.5in]{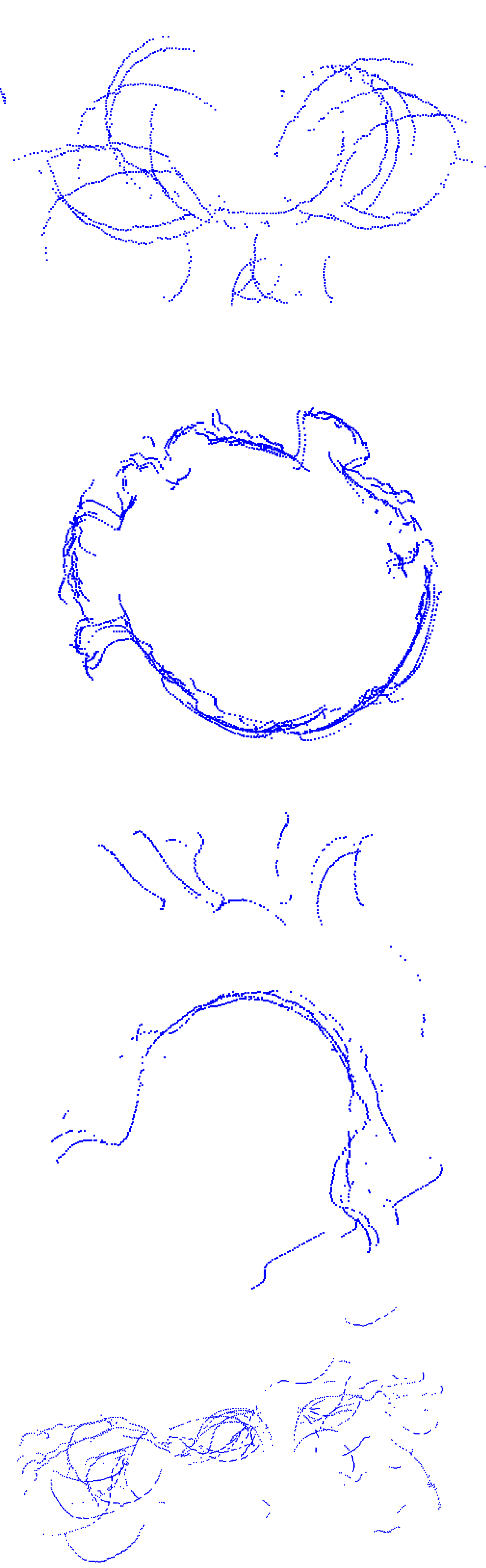} 
        \centerline{(d)}
    \end{minipage}
    }
    \subfloat
    {  \begin{minipage}{0.135\linewidth}
        \flushleft
        \includegraphics[height=3.5in]{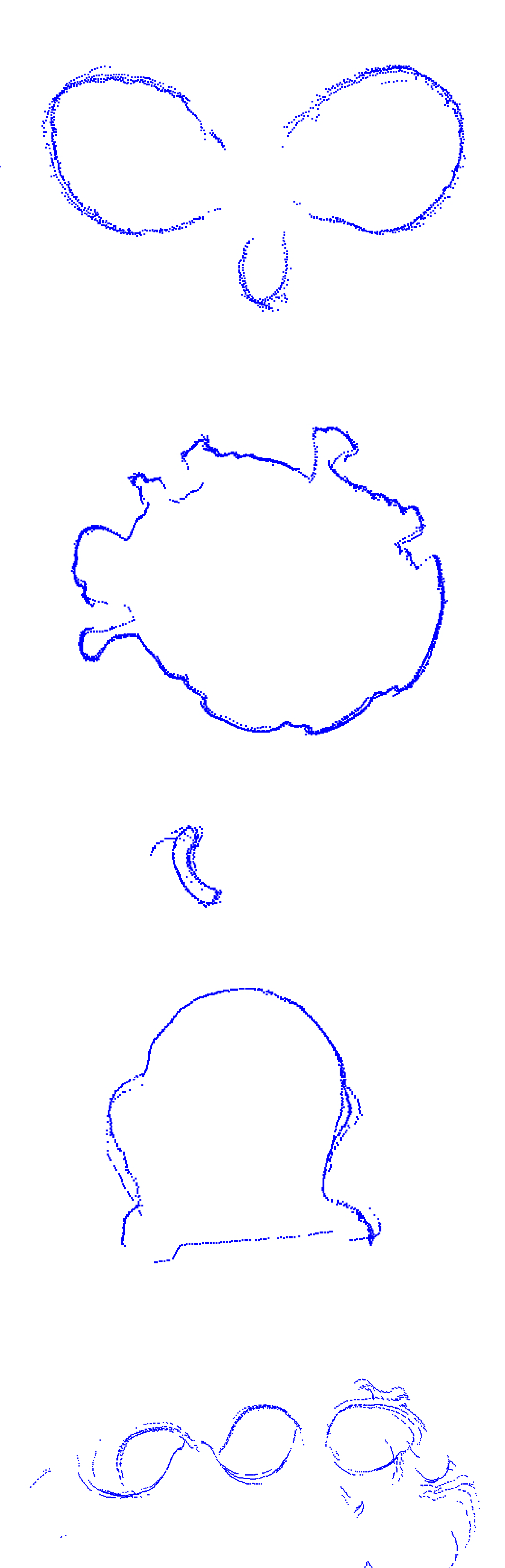}
        \centerline{(e)} 
    \end{minipage}
    }
    \subfloat
    {  \begin{minipage}{0.13\linewidth}
        \flushleft
        \includegraphics[height=3.5in]{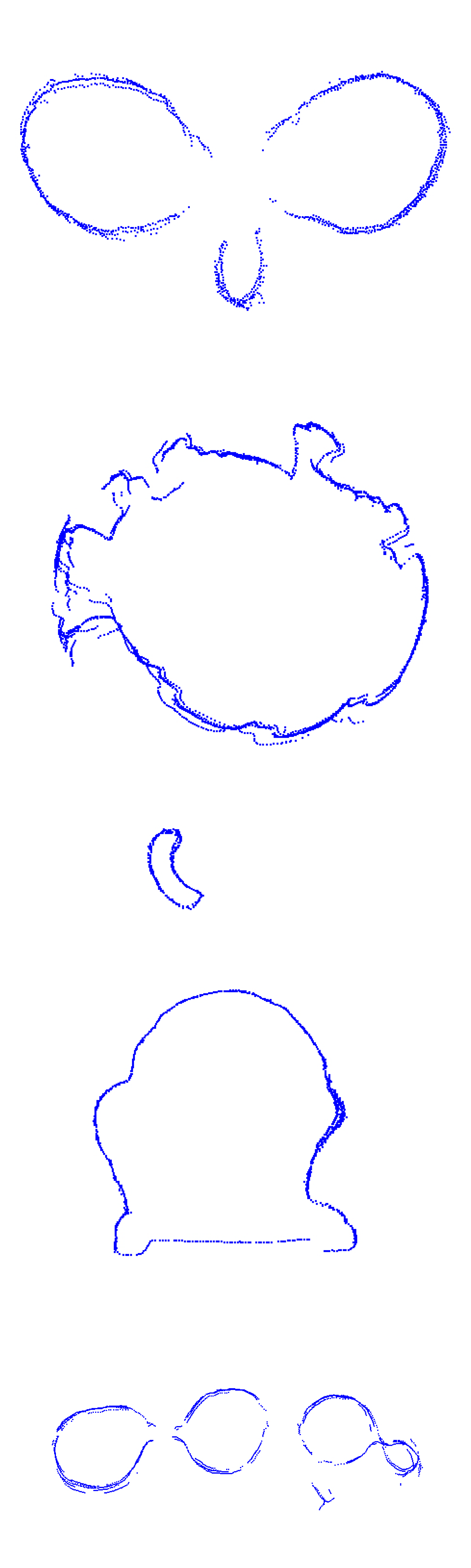} 
        \centerline{(f)}
    \end{minipage}
    }
    \subfloat
    {  \begin{minipage}{0.13\linewidth}
        \flushleft
        \includegraphics[height=3.5in]{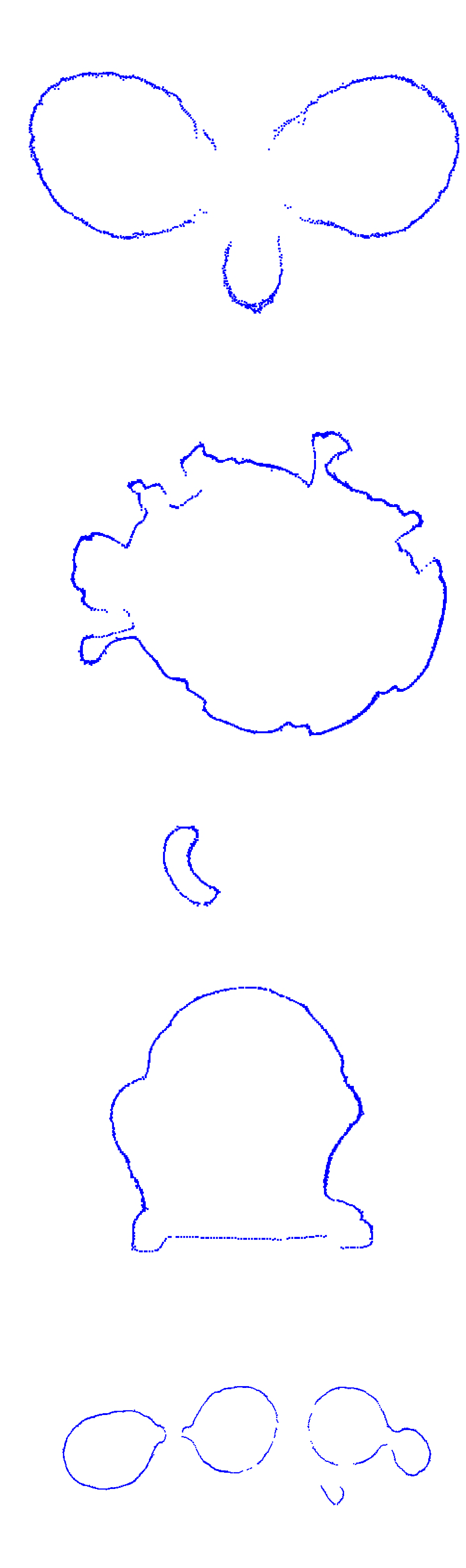} 
        \centerline{   (g)}
    \end{minipage}
    }
    \caption{
      Multi-view registration results of different methods presented in the form of cross-section for 
      four data sets under $20\%$ overlap percentage threshold: (a) Ground truth models. (b) Initial results. (c) LRS results. (d) MA results. 
      (e) WMAA results. (f) MCC-MA results. (g) Our results.
    }
    \label{fig9}
    \end{figure*}

Furthermore, we can see that our performance under $20\%$ overlapping percentage even outperforms the performance under 
$30\%$ overlapping percentage. The fact can be easily interpreted by virtue of the conclusions drawn in Subsection IV.A(1) and IV.A(2): 
the decrease of overlapping percentage leads to the increase of the relevance of each scan, but also the increase of outliers; 
as we saw in Subsection IV.A(1) and IV.A(2), the former will lead to an increase of accuracies of all methods, while the latter will cause only 
a slight decrease the accuracy of our method, but a rapid decrease of those of others methods. This also reveals the reason 
why the registration results achieved by WMAA are not promising, that is, high overlap does not imply a high accuracy of pair-wise 
registration results. In short, one advantage of our method lies in the fact that it can efficiently and accurately exploit the 
redundancy information in the set of relative motions, even though the outlier ratio of the set is relatively high.

\section{Conclusion and Future Work}
In conclusion, we proposed a novel motion averaging framework with Laplacian kernel-based maximum correntropy criterion. 
In particular, we adopted an iterative strategy for determining the global motions similar to many works in previous literature. 
The task at hand, namely, finding the increment of global motions, can then be formulated as an optimization problem aimed at 
maximizing an objective function by virtue of LMCC. Furthermore, the optimization problem can be divided into subproblems: 
obtain the weights according to current residuals and solve a SOCP problem based on the weights, so that our method can efficiently 
avoid the computation of the inverse of a large matrix occurring in many works of literature. We also introduced a new method to determine 
kernel width. Experimental tests on synthetic and data sets demonstrate that our method has state-of-the-art multi-view 
registration accuracy, high robustness to outliers, and low sensibility to initial global motions.

\vspace{11pt}


\vfill

\end{document}